\documentclass{article} 
\usepackage{iclr2022_conference,times}

\usepackage{amsmath,amsfonts,bm}

\def\eqref#1{equation~\ref{#1}}

\def\1{\bm{1}}

\def\va{{\bm{a}}}

\def\ve{{\bm{e}}}

\def\vo{{\bm{o}}}

\def\vs{{\bm{s}}}

\def\evc{{c}}

\def\evl{{l}}

\def\evw{{w}}

\def\mB{{\bm{B}}}

\def\mL{{\bm{L}}}

\def\mX{{\bm{X}}}

\DeclareMathAlphabet{\mathsfit}{\encodingdefault}{\sfdefault}{m}{sl}
\SetMathAlphabet{\mathsfit}{bold}{\encodingdefault}{\sfdefault}{bx}{n}

\def\sC{{\mathbb{C}}}

\def\sL{{\mathbb{L}}}

\def\sS{{\mathbb{S}}}

\def\sW{{\mathbb{W}}}

\newcommand{\R}{\mathbb{R}}

\DeclareMathOperator*{\argmax}{arg\,max}

\usepackage{hyperref}
\usepackage{url}

\usepackage{algorithm}
\usepackage[noend]{algpseudocode}
\usepackage{wrapfig}
\usepackage{tikz}
\usepackage{pgfplots}
\usepackage{pgfplotstable}
\pgfplotsset{compat=newest}
\usepgfplotslibrary{fillbetween}
\usepackage{colortbl}
\usepackage{array}
\usepackage[font=footnotesize]{caption}
\usepackage{longtable}
\usepackage{mathtools}

\definecolor{c0}{RGB}{178,24,43}
\definecolor{c1}{RGB}{239,138,98}
\definecolor{c2}{RGB}{253,219,199}
\definecolor{c3}{RGB}{209,229,240}
\definecolor{c4}{RGB}{103,169,207}
\definecolor{c5}{RGB}{33,102,172}

\definecolor{g76}{RGB}{118,42,131}
\definecolor{g75}{RGB}{175,141,195}
\definecolor{g72}{RGB}{217,240,211}
\definecolor{g71}{RGB}{127,191,123}
\definecolor{g70}{RGB}{27,120,55}

\definecolor{r32}{RGB}{252,146,114}
\definecolor{r31}{RGB}{239,59,44}
\definecolor{r30}{RGB}{153,0,13}

\definecolor{r43}{RGB}{254,229,217}
\definecolor{r42}{RGB}{252,174,145}
\definecolor{r41}{RGB}{251,106,74}
\definecolor{r40}{RGB}{203,24,29}

\definecolor{b32}{RGB}{158,202,225}
\definecolor{b31}{RGB}{66,146,198}
\definecolor{b30}{RGB}{8,69,148}

\definecolor{b43}{RGB}{239,243,255}
\definecolor{b42}{RGB}{189,215,231}
\definecolor{b41}{RGB}{107,174,214}
\definecolor{b40}{RGB}{33,113,181}

\definecolor{g32}{RGB}{161,217,155}
\definecolor{g31}{RGB}{65,171,93}
\definecolor{g30}{RGB}{0,90,50}

\definecolor{v32}{RGB}{239,237,245}
\definecolor{v31}{RGB}{188,189,220}
\definecolor{v30}{RGB}{117,107,177}

\newcolumntype{P}[1]{>{\centering\arraybackslash}p{#1}}
\newcolumntype{M}[1]{>{\centering\arraybackslash}m{#1}}

\newlength{\plotwidth}
\newlength{\plotheight}

\newcommand{\correlationplot}[7]{}
\newcommand{\intersectionplot}[3]{}
\newcommand{\evalplot}[3]{}
\newcommand{\corrlegend}{}
\newcommand{\toplegend}{}
\newcommand{\evallegend}{}
\newcommand{\ablationlegend}{}

\newcommand{\stdopacity}{0.2}
\newcommand{\lineopacity}{1.0}
\newcommand{\legendtextsize}{\scriptsize}

\pgfplotsset{
    every axis/.append style={
        legend style={font=\legendtextsize}
    }
}

\title{Fooling Explanations in Text Classifiers}

\author{Adam Ivankay\\
IBM Research Zurich\\
R\"{u}schlikon, Switzerland \\
\texttt{aiv@zurich.ibm.com} \\
\And
Ivan Girardi\\
IBM Research Zurich\\
R\"{u}schlikon, Switzerland \\
\texttt{ivg@zurich.ibm.com} \\
\And
Chiara Marchiori \\
IBM Research Zurich\\
R\"{u}schlikon, Switzerland \\
\texttt{chi@zurich.ibm.com} \\
\And
Pascal Frossard \\
\'{E}cole Polytechnique F\'{e}d\'{e}rale de Lausanne (EPFL)\\
Lausanne, Switzerland \\
\texttt{pascal.frossard@epfl.ch} \\
}

\iclrfinalcopy
\begin{document}

\maketitle

\begin{abstract}
State-of-the-art text classification models are becoming increasingly reliant on deep neural networks (DNNs). Due to their black-box nature, faithful and robust explanation methods need to accompany classifiers for deployment in real-life scenarios. However, it has been shown in vision applications that explanation methods are susceptible to local, imperceptible perturbations that can significantly alter the explanations without changing the predicted classes. We show here that the existence of such perturbations extends to text classifiers as well. Specifically, we introduce \textsc{TextExplanationFooler (TEF)}, a novel explanation attack algorithm that alters text input samples imperceptibly so that the outcome of widely-used explanation methods changes considerably while leaving classifier predictions unchanged. We evaluate the performance of the attribution robustness estimation performance in TEF on five sequence classification datasets, utilizing three DNN architectures and three transformer architectures for each dataset. TEF can significantly decrease the correlation between unchanged and perturbed input attributions, which shows that all models and explanation methods are susceptible to TEF perturbations. Moreover, we evaluate how the perturbations transfer to other model architectures and attribution methods, and show that TEF perturbations are also effective in scenarios where the target model and explanation method are unknown. Finally, we introduce a \textit{semi-universal} attack that is able to compute fast, computationally light perturbations with no knowledge of the attacked classifier nor explanation method. Overall, our work shows that explanations in text classifiers are very fragile and users need to carefully address their robustness before relying on them in critical applications.
\end{abstract}

\section{Introduction}
\label{sec:intro}
\newcommand{\cellwidth}{0.43\linewidth}

\begin{figure}[t]
\centering
\begin{tabular}{ p{\cellwidth} | p{\cellwidth} | c  }
\centering\textbf{Original sample} & \centering\textbf{TEF sample}  & \textbf{PCC}  \\\hline
\input{figures/examples/sample_files/samples_agnews}                            \\\hline%
\input{figures/examples/sample_files/samples_mr}                                \\%
\end{tabular}

\caption{Example of fragile attributions. Highlighted red words are deemed most important \textit{towards} the predicted class by the Integrated Gradients attribution method, blue ones \textit{against} it. By substituting a few words in the original sample, the \textit{Pearson Correlation Coefficient} (PCC) of word importances drops to below 0.2 while maintaining the same confidence in the correctly predicted class (denoted by $\displaystyle F$).}\label{fig:examples}
\end{figure}
\vspace{\baselineskip}%
Deep neural networks (DNNs) have undoubtedly become the state-of-the-art architectures for many existing machine learning tasks \citep{attentionhealthcare}. Yet, their \textit{black-box} nature has raised the need for developing methods to mitigate the lack of interpretability caused by their increased complexity \citep{saliencymap, occlusion, counterfactuals, attentionmechanism}. These methods give intuitive, easily understandable explanations that do not require significant domain knowledge. This is especially desired in safety-critical scenarios, such as healthcare, where the users of such DNNs - the medical professionals for instance - need to understand the decision process and reasoning behind it. However, they have been shown to lack local robustness towards carefully crafted, imperceptible perturbations in the input \citep{interpretationfragile}. While resulting in the same predictions, these altered inputs yield significantly different explanations and attributions maps (Figure \ref{fig:examples}). Interpretation methods fragile towards small input perturbations not only fail to provide faithful explanations, a desiderata commonly required in explainable AI \citep{faithfulness}, but also damages user trust in DNNs and prevents them from being deployed on high-stakes, safety-critical applications, such as in healthcare \citep{aihealthcare}.\\%
The previously described phenomenon has been widely studied in the image domain by \cite{alignmentconnection, robustnesscurvature} or \cite{far}. However, in discrete-input domains like text, there has been limited progress. This is especially problematic given the increased reliance on and fragility of attention mechanisms as \textit{inherently explainable} methods, as stated in \cite{attentioninherent}. Therefore, we summarize our contributions as follows:
\begin{itemize}
    \item We provide a novel baseline black-box adversarial attack, \textsc{TextExplanationFooler (TEF)} to estimate the local robustness of \textit{explanations} in text classification problems
    \item We evaluate attribution robustness on widely used, state-of-the-art text datasets and model architectures, showing that explanation methods' output can be significantly altered with our attack
    \item We provide insight into the \textit{transfer capability} of \textsc{TEF} on different models and explanation methods as well as introduce \textit{semi-universal} adversarial perturbations to alter explanations without requiring access to the model at attack-time
\end{itemize}

\section{Preliminaries}
\label{sec:prelim}
\subsection{Related Work}%
Adversarial attacks that alter the inference outcomes in DNNs have been widely studied both in the image \citep{adversarialattacks, carliniwagnerattack, deepfool, sparsefool} and text domain \citep{hotflip, advbert, isbertrobust, greedygumbel}. Methods to alleviate the networks susceptibility to such attacks have also been proposed, including the works of \cite{advtraining, curvatureregularization, thermometerencoding} or \cite{parsevalnetworks}. However, it has recently been shown by authors \cite{interpretationfragile} that, in addition to DNN predictions, widely-used \textit{explanation methods} also lack robustness to targeted, imperceptible alterations of the input. These attacks change the outcomes of such explanation methods significantly, while predictions of the DNNs are unaltered. This violates the \textit{Prediction Assumption} of \textit{faithful} explanations and crucially degrades user trust in such explanation methods, as significantly different interpretations are provided for similar inputs and outputs \citep{faithfulness}. The aforementioned phenomenon of fragile explanations has mostly been investigated in the image domain \citep{alignmentconnection, alignment, igsum, far}, with less focus on discrete input spaces like text. However, faithful and robust interpretations are arguably equally important in the discrete text domain, for instance in electronic health record classification \citep{aida} or precision medicine \citep{precisionmedicine}, where critical decisions often need to be based on DNN explanations. The work of \cite{guaranteednlprobustness} constructs inherently robust explanations for NLP models, however only towards perturbations in the embedding space, not the input space that adversaries can operate on. Moreover, they do not give a method to evaluate robustness of already existing explanation algorithms. The authors of \cite{nnpathologies} show that interpretation methods in NLP lack completeness \citep{integratedgradients} by removing words deemed least important by explanation methods. Moreover, attention mechanisms \citep{attentionmechanism} have been increasingly relied on as \textit{inherently} interpretable systems. However recent work has questioned their faithfulness and plausibility \citep{attentioninterpretable, attentionnotexplanation, attentionnotnotexplanation} by proving that they often do not highlight input components that are most important to a DNN decision. We, paralleled by the very recent work of \cite{devilswork}, are the first to show that imperceptible perturbations in the \textit{input space} can alter the outcome of explanations of text classifiers significantly, giving an efficient attack to estimate explanation robustness. However, our work is the first to give an extensive evaluation of the robustness of widely-used explanation methods on large datasets, comparing several state-of-the-art architectures. Further, we are the first to address robustness of attention weights in transformer architectures and to provide insight on transfer capabilities and universal attacks.
\subsection{Background}%
Let $\displaystyle \sS = \displaystyle \{\displaystyle \vs_1, \displaystyle \vs_2, ..., \displaystyle \vs_N\}$ be a dataset of $N$ text samples $\displaystyle \vs_i$, each with a label from a predefined set of labels $\displaystyle \sL = \displaystyle \{\displaystyle \evl_1, \displaystyle \evl_2, ..., \displaystyle \evl_{|\sL|}\}$. Each sample $\displaystyle \vs_i$ contains a sequence of tokens (or words) $\displaystyle \evw_i$ taken from a discrete vocabulary set $\displaystyle \sW = \displaystyle \{\displaystyle \evw_1, \displaystyle \evw_2, ..., \displaystyle \evw_{|\sW|}\}$. A generic sequence classifier then consist of a non-injective, non-surjective embedding function $\displaystyle E: \sS \rightarrow \R^{d \times p} ,\; \displaystyle E(\vs) = \mX$, which maps the input sample $\displaystyle \vs$ to its embedding matrix $\displaystyle \mX$, and a function $\displaystyle f: \R^{d \times p} \rightarrow \R^{|\sL|},\; \displaystyle f(\displaystyle \mX) = \displaystyle \vo$, representing a (DNN) classifier function. $d$ and $p$ denote the embedding dimension and sequence length respectively. Let $\displaystyle F: \sS \rightarrow \R^{|\sL|},\; \displaystyle F(\vs) = f \circ E$ be the full sequence classifier with final prediction $\displaystyle y = {\argmax}_{i \in \{1: {|\mL|}\}}{\vo_i}$.\\%
We define an attribution map as $\displaystyle A: \sS \rightarrow \R^p,\; \displaystyle A(\displaystyle \vs, F, \evl) = \displaystyle \va$ that assigns a scalar value to each input token $\displaystyle \evw_i$ in the text sample $\displaystyle \vs$, resulting in the attribution vector $\displaystyle \va \in \R^p$. This vector represents each token's influence towards the prediction outcome $\displaystyle y$ of classifier $\displaystyle F$. Our work considers three widely-used attribution methods in text classification, namely Saliency Maps (S) \citep{saliencymap}, Integrated Gradients (IG) \citep{integratedgradients} and Attention (A) \citep{attentionmechanism}, defined in the following Equations (\ref{eqn:saliencymap}), (\ref{eqn:intgrad}) and (\ref{eqn:attention}) respectively.%
\begin{equation}
    \label{eqn:saliencymap}
    A^{\mathrm{S}}_i(\vs, F, l) {}={} \sum_{j \in \{1: d\}} |\nabla_\mX f(\mX)_l |_{j, i}
\end{equation}
\begin{equation}
\label{eqn:intgrad}
    A^{\mathrm{IG}}_i(\vs, F, l, \mB) {}={} \sum_{j \in \{1: d\}} \big[ (\mX - \mB) \cdot \int_{\alpha=0}^{1} \nabla_{\tilde{\mX}} f(\tilde{\mX})_l |_{\tilde{\mX} {}={} \mB + \alpha(\mX - \mB)}\, d\alpha \big ]_{j, i}
\end{equation}
\begin{equation}
    \label{eqn:attention}
    A^{\mathrm{Att}}_i(\vs, F, l) {}={} \frac{\exp{e_{i}}}{\sum_{j \in \{1: p\}}\exp{e_{j}}}
\end{equation}
where $\displaystyle \mB$ denotes the null matrix $\mathbf{0}^{d \times p}$, $f$ is the classifier function of $F$, $\alpha$ a scaling factor and $\displaystyle \mX {} = {} E(\vs)$. $\nabla_\mX f$ denotes the matrix-derivative of $f$ to $\mX$, as defined in \cite{deeplearningbook}. An attention head is a layer that transforms its inputs into scores $\displaystyle \ve$ and calculates its output by linear combination of each input score, with coefficients normalized to a distribution. These coefficients are the attention weights $A^{\mathrm{Att}}_i(\vs, F, l)$ denoted in Equation (\ref{eqn:attention}). It is commonly agreed to give intuitive explanations on how much the model \textit{attends} to the given inputs through its attention weights \citep{faithfulness}.%

\section{Methods}
\label{sec:methods}
In this section, we describe our novel method \textsc{TextExplanationFooler (TEF) }to estimate attribution robustness (AR) in sequence classification problems. Specifically, we define the problem formulation, introduce our threat model and present the algorithm used to alter explanations by imperceptibly changing the inputs.%
\subsection{Problem Formulation}%
Given an input text samples $\displaystyle \vs$ and $\displaystyle \tilde{\vs}$, labels $\displaystyle \evl$; a text classifier $F$ with embedding function $E$ and classifier function $f$; and attribution method $A$, we define \textbf{attribution robustness} (also \textit{explanation robustness}, AR) as written in Equation (\ref{eqn:attributionrobustness}).%
\begin{equation}
    \label{eqn:attributionrobustness}
    r(\tilde{\va}, \va) {}={} 1 {}-{} \max_{\tilde{\va}} \; d(\tilde{\va},\va) {}={} 1 {}-{} \max_{\tilde{\vs}} \; d \big[ A(\tilde{\vs}, F, l), \; A(\vs, F, l) \big]
\end{equation}
with
\begin{equation}
\label{eqn:predictionconstraint}
    \argmax_{i \in \{1:|\sL|\}}F(\tilde{\vs}) {}={} \argmax_{i \in \{1:|\sL|\}}F(\vs),
\end{equation}
where $d$ denotes a distance measure between the attributions $\displaystyle \tilde{\va}$ and $\displaystyle \va$ of the the two input samples $\displaystyle \vs$ and $\displaystyle \tilde{\vs}$. The rest of the notation is kept as in Section \ref{sec:prelim}. Equation (\ref{eqn:attributionrobustness}) quantifies how different the attributions of two input samples are, given the constraint in Equation (\ref{eqn:predictionconstraint}) that enforces the inputs having the same prediction outcome.\\%
The attribution robustness estimation is then solved utilizing the following Equation (\ref{eqn:optimizationobjective}).%
\begin{equation}
\label{eqn:optimizationobjective}
    \vs_{\mathrm{adv}} {}={} \argmax_{\tilde{\vs}} \; d \big[ A(\tilde{\vs}, F, l), \; A(\vs, F, l) \big]
\end{equation}
where $\displaystyle \vs_{\mathrm{adv}}$ denotes the solution to the estimation, i.e. the adversarial input, which also minimizes AR defined in Equation (\ref{eqn:attributionrobustness}). $\displaystyle \vs$ denotes the original, unperturbed input and $\displaystyle \tilde{\vs}$ the perturbed input, optimized during estimation. The solution $\displaystyle \vs_{\mathrm{adv}}$ gives a robustness estimate by finding an input that maximizes the distance between original attribution $\displaystyle A(\vs, F, l)$ and adversarial attribution $A(\tilde{\vs}, F, l)$ within a local neighbourhood of $\displaystyle \vs$. The more \textit{dissimilar} these maps are, the less robust the attribution method is. The local neighbourhood is defined by both linguistic constraints described in the next section that encourage semantic proximity to the original text and the perturbed samples having the same prediction outcome as the unperturbed ones, see Equation (\ref{eqn:predictionconstraint}). This formulation is backed by current research \citep{interpretationfragile, geometryblame, far} and the \textit{Prediction Assumption} of faithful explanations \citep{faithfulness}. 
\subsection{Threat Model and Attack}%
\begin{wrapfigure}{R}{0.55\textwidth}%
\vspace{-2\baselineskip}%
\begin{minipage}{0.55\textwidth}%
\input{algorithms/tef}%
\end{minipage}%
\vspace{-3\baselineskip}%
\end{wrapfigure}%
We define our algorithm to estimate AR as a \textit{black-box} attack. It only queries the model to obtain its output logits and the accompanied explanations of the inference process. The model might access its gradients to compute explanations, but the attack only utilizes the resulting explanations, no gradient or architectural information. We restrict the valid input perturbations to token substitutions, specifically insertions and deletions of tokens are forbidden, as they alter the input lengths. Algorithm \ref{alg:tef} contains the schematic code for \textsc{TEF}, consisting of the following two steps.%
\paragraph{Step 1 - Word importance ranking} First, an importance ranking is extracted for each token of the input sample. Specifically, we compute $\displaystyle I_{\evw_i} {}={} d \big[ A(\vs_{w_i\rightarrow 0}, F, l), \; A(\vs, F, l) \big]$ for each token $i$ in $\displaystyle \vs$, where $\displaystyle \vs_{w_i\rightarrow 0}$ denotes the input sequence $\displaystyle \vs$ with the $i$-th word masked to the zero embedding token. The input tokens are then sorted by the $\displaystyle I_{\evw_i}$ values in a decreasing fashion. Then, high importance words are prioritized during substitution.%
\paragraph{Step 2 - Candidate selection} For each word $\displaystyle w_i$ in $\displaystyle \vs$ sequentially, a set of \textit{substitution candidates} $\displaystyle \sC$ of $N$ elements is extracted. This candidate set is constructed from the counter-fitted GloVe \citep{glove} synonym embeddings by the authors of \cite{counterfitted}. The candidates are then filtered by Part-Of-Speech (POS), tagged by SpaCy \citep{spacy}, only allowing replacements with equal POS. Stop words are also discarded from $\sC$. A \textit{final selection} as replacement for $\displaystyle w_i$ is then made to be the $\displaystyle \evc_k \in \sC$ that maximizes $\displaystyle d \big[ A(\tilde{\vs}_{w_i\rightarrow \evc_k}, F, l), \; A(\vs, F, l) \big]$. The algorithm is aborted when the number of replacements to sentence length exceeds the maximum value $\displaystyle \rho_{max}$.%
%
%

\section{Experiments and Results}
\label{sec:exp_res}
In this section, we present an extensive evaluation of our attribution robustness (AR) estimation attack, \textsc{TEF}, for sequence classification problems. We examine the performance of TEF and study the impact of different factors on its robustness evaluation performance. We find that our attack effectively reduces the correlation of original and attacked attributions on all datasets and models. Moreover, we describe our transfer and semi-universal attacks and examine their robustness estimation performance, showing that even under circumstances where the model and explainer are unknown to the attacker, TEF perturbations transferred from other models decrease attribution robustness effectively.%
\subsection{Models, datasets and evaluation}%
%
%
\begin{table}[t]

\vspace{\baselineskip}

\caption{Accuracies, average text length and number of classes of our models trained on the five datasets.}\label{tbl:modelsdatasets}

\vspace{\baselineskip}

\centering

\addtolength{\tabcolsep}{-2pt}

\begin{tabular}{c|cccccc|c|c}

\textsc{\textbf{Dataset}}   & \textsc{CNN}     & \textsc{LSTM}    & \textsc{LSTMAtt}   & \textsc{BERT}      & \textsc{RoBERTa}   & \textsc{XLNet}     & Mean $|\vs|$  & $|\sL|$   \\\hline
\textsc{AG's News}          & 89.7\%  & 90.8\%  & 91.4\%    & 94.2\%    & 94.0\%    & 93.8\%    & 45            & 4         \\
\textsc{IMDB}               & 82.0\%  & 87.2\%  & 87.3\%    & 89.4\%    & 93.3\%    & 93.7\%    & 270           & 2         \\
\textsc{Fake News}          & 98.9\%  & 99.6\%  & 99.6\%    & 99.8\%    & 100.0\%   & 100.0\%   & 919           & 2         \\
\textsc{MR}                 & 73.0\%  & 76.4\%  & 78.0\%    & 82.2\%    & 87.7\%    & 86.3\%    & 22            & 2         \\
\textsc{Yelp}               & 49.0\%  & 54.8\%  & 60.0\%    & 62.6\%    & 67.6\%    & -    & 159           & 5         \\

\end{tabular}

\addtolength{\tabcolsep}{2pt}

\end{table}
\vspace{\baselineskip}%
\renewcommand{\correlationplot}[3]{

    \begin{tikzpicture}
    
	\pgfplotstableread{data_new/tef/#1/#2/#3_auc.tex}{\accteftable}
	\pgfplotstablegetelem{0}{[index]0}\of\accteftable
	\pgfmathsetmacro{\acctef}{\pgfplotsretval}

	\pgfplotstableread{data_new/tef/#1/#2/#3.tex}{\teftable}
	\pgfplotstablegetrowsof{\teftable} 
	\pgfmathsetmacro{\teflastrow}{\pgfplotsretval-1}
	
	\pgfplotstablegetelem{\teflastrow}{expl_sim_mean}\of\teftable
	\pgfmathsetmacro{\maxtef}{\pgfplotsretval}
	
	\pgfplotstableread{data_new/ra/#1/#2/#3_auc.tex}{\accratable}
	\pgfplotstablegetelem{0}{[index]0}\of\accratable
	\pgfmathsetmacro{\accra}{\pgfplotsretval}

	\pgfplotstableread{data_new/ra/#1/#2/#3.tex}{\ratable}
	\pgfplotstablegetrowsof{\ratable} 
	\pgfmathsetmacro{\ralastrow}{\pgfplotsretval-1}
	
	\pgfplotstablegetelem{\ralastrow}{expl_sim_mean}\of\ratable
	\pgfmathsetmacro{\maxra}{\pgfplotsretval}

    \begin{axis}[
      width=\plotwidth,
      height=\plotheight,
      xmin=0.0, xmax=0.32,
      ymin=-0.6, ymax=1,
      ymajorgrids=true,
      xmajorgrids=true,
      ytick={-0.5, 0, 0.5 ,1},
      yticklabels={, 0, , 1},
      xtick={0.16, 0.32},
      xticklabels={0.16, $\displaystyle \rho$},
      grid style=dashed
    ]

    
    \addplot[name path=kendall_mean, r32, mark=triangle, line width=1pt, opacity=\lineopacity] table[x={x}, y={kendall_mean}] {data_new/tef/#1/#2/#3.tex};\label{pgf:corrkendalltef};
    \addplot[name path=kendall_mean, dashed, r32, mark=triangle, mark options={solid}, line width=1pt, opacity=\lineopacity] table[x={x}, y={kendall_mean}] {data_new/ra/#1/#2/#3.tex};\label{pgf:corrkendallra};
%
%
    \addplot[name path=spearman_mean, r31, mark=diamond, line width=1pt, opacity=\lineopacity] table[x={x}, y={spearman_mean}] {data_new/tef/#1/#2/#3.tex};\label{pgf:corrspearmantef};
    \addplot[name path=spearman_mean, dashed, r31, mark=diamond, mark options={solid}, line width=1pt, opacity=\lineopacity] table[x={x}, y={spearman_mean}] {data_new/ra/#1/#2/#3.tex};\label{pgf:corrspearmanra};
%
%
    \addplot[name path=expl_sim_mean, r30, mark=*, line width=1pt, opacity=\lineopacity] table[x={x}, y={expl_sim_mean}] {data_new/tef/#1/#2/#3.tex};\label{pgf:corrpcctef};
    \addplot[name path=expl_sim_mean, dashed, mark options={solid}, r30, mark=o, line width=1pt, opacity=\lineopacity] table[x={x}, y={expl_sim_mean}] {data_new//ra/#1/#2/#3.tex};\label{pgf:corrpccra};

    
    \addplot[name path=klq, r32!50, opacity=\stdopacity] table[x={x}, y={kendall_lower_quartile}] {data_new/tef/#1/#2/#3.tex};
    \addplot[name path=kuq, r32!50, opacity=\stdopacity] table[x={x}, y={kendall_upper_quartile}] {data_new/tef/#1/#2/#3.tex};
    \addplot[r32!50,fill opacity=\stdopacity] fill between[of=klq and kuq];

    \addplot[name path=slq, r31!50, opacity=\stdopacity] table[x={x}, y={spearman_lower_quartile}] {data_new/tef/#1/#2/#3.tex};
    \addplot[name path=suq, r31!50, opacity=\stdopacity] table[x={x}, y={spearman_upper_quartile}] {data_new/tef/#1/#2/#3.tex};
    \addplot[r31!50,fill opacity=\stdopacity] fill between[of=slq and suq];
    
    \addplot[name path=plq, r30!50, opacity=\stdopacity] table[x={x}, y={expl_sim_lower_quartile}] {data_new/tef/#1/#2/#3.tex};
    \addplot[name path=puq, r30!50, opacity=\stdopacity] table[x={x}, y={expl_sim_upper_quartile}] {data_new/tef/#1/#2/#3.tex};
    \addplot[r30!50,fill opacity=\stdopacity] fill between[of=plq and puq];

	\node at (axis cs:0.32, \maxtef) [anchor=north east, opacity=\lineopacity] {\scriptsize\textbf{\textcolor{r30}{ACC}: \acctef}};
	\node at (axis cs:0.32, \maxra) [anchor=north east, opacity=\lineopacity] {\scriptsize\textbf{\textcolor{r30}{ACC}: \accra}};

    \end{axis}
\end{tikzpicture}
}

\renewcommand{\intersectionplot}[3]{

    \begin{tikzpicture}
        \begin{axis}[
          width=\plotwidth,
          height=\plotheight,
          xmin=0.0, xmax=0.32,
          ymin=-0.6, ymax=1,
          ymajorgrids=true,
          xmajorgrids=true,
          yticklabels=\empty,
          xtick={0.16, 0.32},
          xticklabels={0.16, $\displaystyle \rho$},
          grid style=dashed
        ]
    
        \addplot[name path=top10_mean, b30, mark=*, line width=1pt, opacity=\lineopacity] table[x={x}, y={top10_mean}] {data_new/tef/#1/#2/#3.tex};\label{pgf:top10tef};
        \addplot[name path=top10_mean, dashed, mark options={solid}, b30, mark=o, line width=1pt, opacity=\lineopacity] table[x={x}, y={top10_mean}] {data_new/ra/#1/#2/#3.tex};\label{pgf:top10ra};
    
        \addplot[name path=top30_mean, b31, line width=1pt, mark=triangle, opacity=\lineopacity] table[x={x}, y={top30_mean}] {data_new/tef/#1/#2/#3.tex};\label{pgf:top30tef};
        \addplot[name path=top30_mean, dashed, mark options={solid}, b31, line width=1pt, mark=triangle, opacity=\lineopacity] table[x={x}, y={top30_mean}] {data_new/ra/#1/#2/#3.tex};\label{pgf:top30ra};
        
        \addplot[name path=top50_mean, b32, mark=diamond, line width=1pt, opacity=\lineopacity] table[x={x}, y={top50_mean}] {data_new/tef/#1/#2/#3.tex};\label{pgf:top50tef};
        \addplot[name path=top50_mean, dashed, mark=diamond, mark options={solid}, b32, line width=1pt, opacity=\lineopacity] table[x={x}, y={top50_mean}] {data_new/ra/#1/#2/#3.tex};\label{pgf:top50ra};
%
%
        \addplot[name path=t10l, b30!50, opacity=\stdopacity] table[x={x}, y={top10_lower_quartile}] {data_new/tef/#1/#2/#3.tex};
        \addplot[name path=t10u, b30!50, opacity=\stdopacity] table[x={x}, y={top10_upper_quartile}] {data_new/tef/#1/#2/#3.tex};
        \addplot[b30!50,fill opacity=\stdopacity] fill between[of=t10l and t10u];
%
%
        \addplot[name path=t30l, b31!50, opacity=\stdopacity] table[x={x}, y={top30_lower_quartile}] {data_new/tef/#1/#2/#3.tex};
        \addplot[name path=t30u, b31!50, opacity=\stdopacity] table[x={x}, y={top30_upper_quartile}] {data_new/tef/#1/#2/#3.tex};
        \addplot[b31!50,fill opacity=\stdopacity] fill between[of=t30l and t30u];
%
%
        \addplot[name path=t50l, b32!50, opacity=\stdopacity] table[x={x}, y={top50_lower_quartile}] {data_new/tef/#1/#2/#3.tex};
        \addplot[name path=t50u, b32!50, opacity=\stdopacity] table[x={x}, y={top50_upper_quartile}] {data_new/tef/#1/#2/#3.tex};
        \addplot[b32!50,fill opacity=\stdopacity] fill between[of=t50l and t50u];
        \end{axis}
    \end{tikzpicture}
}

\renewcommand{\evalplot}[3]{

    \begin{tikzpicture}

	\pgfplotstableread{data_new/tef/#1/#2/#3_ppl.tex}{\ppltable}
	\pgfplotstablegetelem{0}{[index]0}\of\ppltable
	\pgfmathsetmacro{\pplmax}{\pgfplotsretval}
    
        \begin{axis}[
          width=\plotwidth,
          height=\plotheight,
          xmin=0.0, xmax=0.32,
          ymin=-0.6, ymax=1,
          ymajorgrids=true,
          xmajorgrids=true,
          ytick={-0.5, 0, 0.5 ,1},
          yticklabels={, 0, , \pplmax},
          yticklabel pos=right,
          xtick={0.16, 0.32},
          xticklabels={0.16, $\displaystyle \rho$},
          grid style=dashed
        ]
    
        \addplot[name path=sem_sim_mean, g30, mark=diamond, line width=1pt, opacity=\lineopacity] table[x={x}, y={sem_sim_mean}] {data_new/tef/#1/#2/#3.tex};\label{pgf:evalss};
        \addplot[name path=ra_rel_ppl_increase, dashed, g31, mark=triangle, line width=1pt, opacity=\lineopacity] table[x={x}, y={rel_ppl_increase}] {data_new/ra/#1/#2/#3.tex};\label{pgf:evalpplra};
        \addplot[name path=tef_rel_ppl_increase, g32, mark=triangle, line width=1pt, opacity=\lineopacity] table[x={x}, y={rel_ppl_increase}] {data_new/tef/#1/#2/#3.tex};\label{pgf:evalppltef};
%
        \addplot[name path=ssl, g30!50, opacity=\stdopacity] table[x={x}, y={sem_sim_lower_quartile}] {data_new/tef/#1/#2/#3.tex};
        \addplot[name path=ssu, g30!50, opacity=\stdopacity] table[x={x}, y={sem_sim_upper_quartile}] {data_new/tef/#1/#2/#3.tex};
        \addplot[g30!50,fill opacity=\stdopacity] fill between[of=ssl and ssu];
     
        \end{axis}
    \end{tikzpicture}
}

\renewcommand{\corrlegend}{
	\begin{tikzpicture}
    	\node[draw,fill=white,inner sep=5pt,above left=0.5em] at (0,0) {
    	\begin{tabular*}{0.64\plotwidth}{c @{\extracolsep{\fill}} cc}
    			& \textbf{TEF}				& \textbf{RA} 				\\
    		PCC & \ref{pgf:corrpcctef}		& \ref{pgf:corrpccra}		\\
    		ROC & \ref{pgf:corrkendalltef}	& \ref{pgf:corrkendallra}	\\
    		SCC & \ref{pgf:corrspearmantef}	& \ref{pgf:corrspearmanra}
    	\end{tabular*}};
\end{tikzpicture}
}

\renewcommand{\toplegend}{
	\begin{tikzpicture}
    	\node[draw,fill=white,inner sep=5pt,above left=0.5em] at (0,0) {
    	\begin{tabular*}{0.65\plotwidth}{c @{\extracolsep{\fill}} cc}
    					& \textbf{TEF}			& \textbf{RA} 		\\
    		Top-10\% 	& \ref{pgf:top10tef}		& \ref{pgf:top10ra}	\\
    		Top-30\% 	& \ref{pgf:top30tef}		& \ref{pgf:top30ra}	\\
    		Top-50\% 	& \ref{pgf:top50tef}		& \ref{pgf:top50ra}
    	\end{tabular*}};
\end{tikzpicture}
}

\renewcommand{\evallegend}{
	\begin{tikzpicture}
    	\node[draw,fill=white,inner sep=5pt,above left=0.5em] at (0,0) {
    	\begin{tabular*}{0.6\plotwidth}{c @{\extracolsep{\fill}} cc}
    					& \textbf{TEF}			& \textbf{RA} 				\\
    		PPL inc. 	& \ref{pgf:evalppltef}	& \ref{pgf:evalpplra}	\\
    		Sem. sim. 	& \multicolumn{2}{c}{\ref{pgf:evalss}}	\\
    	\end{tabular*}
    	};
\end{tikzpicture}
}

\addtolength{\tabcolsep}{-8pt}  

\setlength{\plotwidth}{0.4\textwidth}
\setlength{\plotheight}{0.25\textwidth}

\begin{figure}[t]
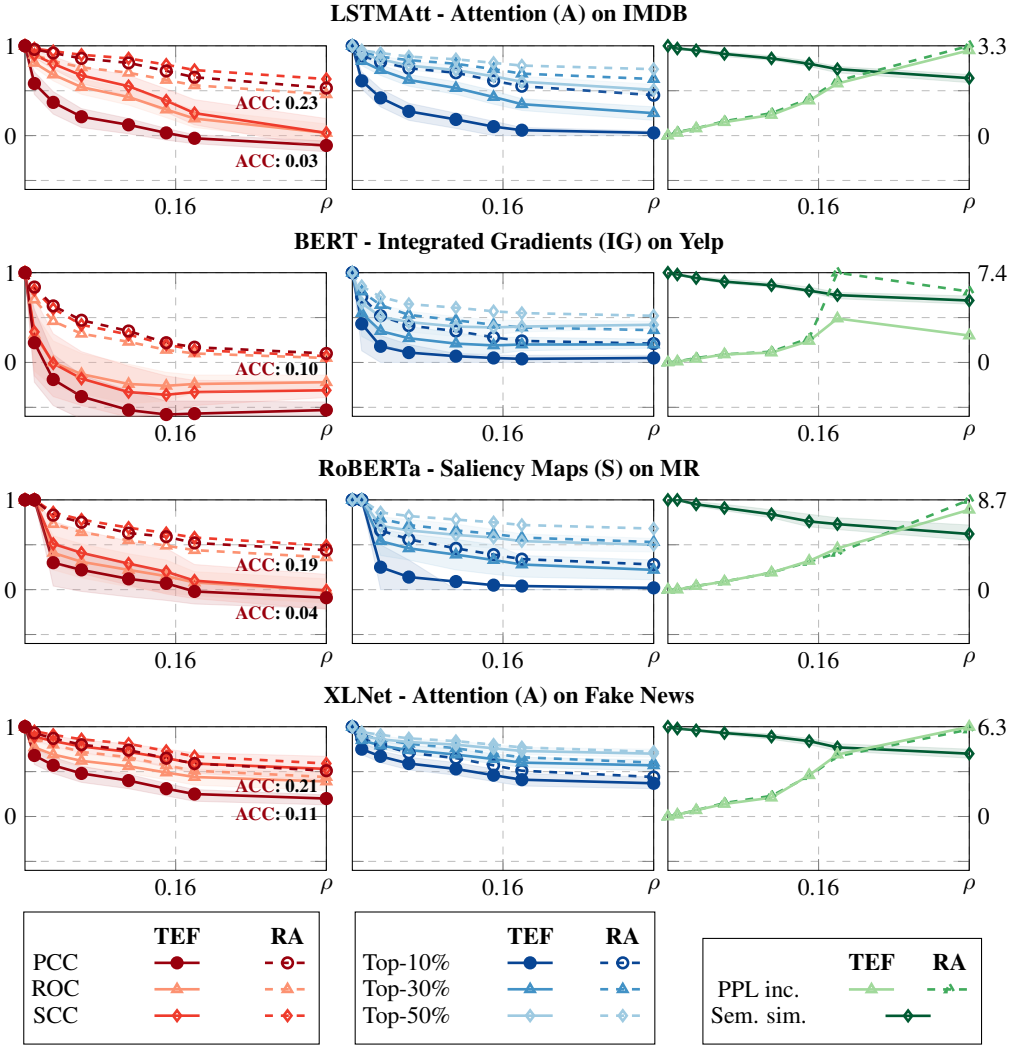

    \centering
    \footnotesize
    \begin{tabular}{ccc}
        \multicolumn{3}{c}{\textbf{LSTMAtt - Attention (A) on IMDB}}	
\\
        \correlationplot{imdb}{lstmatt}{A}
		& \intersectionplot{imdb}{lstmatt}{A}
		& \evalplot{imdb}{lstmatt}{A}
\\
        \multicolumn{3}{c}{\textbf{BERT - Integrated Gradients (IG) on Yelp}}
\\
        \correlationplot{yelp}{bert}{IG}
        & \intersectionplot{yelp}{bert}{IG}
        & \evalplot{yelp}{bert}{IG}
\\
        \multicolumn{3}{c}{\textbf{RoBERTa - Saliency Maps (S) on MR}}
\\
        \correlationplot{mr}{roberta}{S}   
        & \intersectionplot{mr}{roberta}{S}
        & \evalplot{mr}{roberta}	{S}
\\
        \multicolumn{3}{c}{\textbf{XLNet - Attention (A) on Fake News}}
\\
        \correlationplot{fakenews}{xlnet}{A}
        & \intersectionplot{fakenews}{xlnet}{A}
        & \evalplot{fakenews}{xlnet}{A}
\\
		\corrlegend
		& \toplegend
		& \evallegend

    \end{tabular}
    \caption{Robustness of attribution maps on several architectures and explainers. We plot the average correlations (PCC, ROC, SCC) (left), the Top-10\%, Top-30\% and Top-50\% intersections (middle), the semantic similarity and increase of average perplexity (right) as functions of the perturbed ratio $\displaystyle \rho$. Dashed lines indicate the metrics for our \textsc{RandomAttack} (RA). The ACC indicates the area under the PCC (\ref{pgf:corrpcctef} and \ref{pgf:corrpccra}) curve, lower values correspond to overall lower feature attribution correlations in the overall operation interval of $\displaystyle \rho$. The perplexity increases are indicated on the right axis, all other metrics on the left.}
    \label{fig:stdevalcurves}
\end{figure}

\addtolength{\tabcolsep}{8pt}  
Our TEF attack is evaluated on five commonly used public sequence classification datasets, AG's News \citep{agnews_mr}, MR reviews \citep{agnews_mr}, IMDB Movie Reviews \citep{imdb}, Fake News Dataset \footnote{https://www.kaggle.com/c/fake-news/data} and Yelp \citep{yelp}. We train six different word embedding-based architectures for each dataset, namely a CNN, an LSTM, an LSTM containing a single attention layer with one head (LSTMAtt) and three state-of-the-art finetuned transformer-based architectures, BERT \citep{bertattack}, RoBERTa \citep{roberta} and XLNet \citep{xlnet}. Table \ref{tbl:modelsdatasets} contains a summary of our model performances as well as details on the datasets. The text samples are tokenized with the default English SpaCy \citep{spacy} tokenizer for the CNN, LSTM and LSTMAtt models and embedded with the pretrained GloVe 6B 300-dimensional word vectors \citep{glove}. The transformer-based models use their own pretrained tokenizers and embeddings. We use PyTorch \citep{pytorch} with Captum \citep{captum} to implement our models and explainers and the Huggingface Transformers library \citep{huggingface} to finetune the transformer architectures on our datasets.\\%
We evaluate the robustness of three commonly used explanation methods in natural language processing with our \textsc{TEF} attack. These are Saliency Maps (S), Integrated Gradients (IG) and the Attention mechanism (A), defined in Section \ref{sec:prelim}. We use S and IG in combination with all our architectures, Attention only with LSTMAtt, BERT, RoBERTa and XLNet. During the attack, we set the attribution distance $d$ of Equation (\ref{eqn:optimizationobjective}) to be $\displaystyle d(\tilde{\va},\va) = 1 - \frac{\mathrm{PCC}(\tilde{\va},\va) + 1}{2}$, with $\mathrm{PCC}$ denoting the Pearson Correlation Coefficient \citep{pcc} of original and adversarial attributions $\displaystyle \tilde{\va}$ and $\displaystyle \va$. We then report the standard Pearson Correlation Coefficient (PCC), Kendall's Rank Order Correlation (ROC) \citep{kendall}, Spearman's Correlation Coefficient (SCC) \citep{spearman} and the Top-10\%, Top-30\% and Top-50\% intersections to measure AR in Equation (\ref{eqn:attributionrobustness}). These are common metrics that correspond to human measures of AR \citep{interpretationfragile, geometryblame, far}. Additionally, in order to quantify imperceptibility of perturbations, the \textit{semantic similarity} of adversarially perturbed and unchanged sentences is reported, along with the relative increase of average \textit{perplexity} of the perturbed samples, given by the GPT-2 \citep{gpt2} language model. Semantic similarity (Sem. sim.) is measured by the cosine distance between the embeddings produced by the Universal Sentence Encoder (USE) \citep{use}. This is a state-of-the-art sentence embedding widely used in adversarial attacks on text \citep{advbert, isbertrobust}. Perplexity increase (PPL inc.) indicates how much the \textit{likelihood} of the perturbed data has decreased, given a language model, and is often used to validate language models \citep{perplexity}.\\%
Due to the lack of related work in this field, we compare the AR estimation performance of TEF to our \textsc{RandomAttack} (RA) baseline. RA serves as an agnostic attack, utilizes a random word importance ranking in Step 1 of TEF and selects a random synonym in the \textit{final selection} in Step 2. POS and stop word filters (see Section \ref{sec:methods}) are still utilized in RA to keep linguistic constraints intact.%

\subsection{Robustness of explanations}%
\paragraph{Attribution robustness estimation.} In order to estimate the attribution robustness (AR) of the aforementioned models and explainers, we vary the parameter $\displaystyle \rho_{max}$ of TEF, which denotes the maximum ratio of perturbed tokens in the input sample. A larger $\displaystyle \rho_{max}$ value leads to lower attribution correlation, as potentially more words are substituted in the input. We then capture the aforementioned metrics PCC, ROC, SCC, Sem. sim, Top-10\%/30\%/50\% intersections and PPL inc. to evaluate AR. Additionally, in order to quantify performance of our attack over the whole operation interval of $\displaystyle 0 \leq \rho_{max} \leq 0.4$, we compute the Area under the Pearson Correlation Curve (ACC). A lower value of ACC corresponds to lower robustness overall, as correlation values are lower. We note that a particular value of $\displaystyle \rho_{max}$ does not guarantee that all input samples have exactly $\displaystyle \rho_{max}$ ratio of perturbed tokens. Therefore, we quantize our samples based on their actual, resulting perturbed ratio $\displaystyle \rho$ such that samples with similar $\displaystyle \rho$ are grouped together. These bins are computed per dataset, ensuring the comparability of resulting curves and ACCs for each plot. Moreover, we choose the number of candidates in Step 2 of TEF to be $N {}={} |\sC| {}={} 15$, as it is a good trade-off between TEF estimation performance and attack run time. As expected, we find that TEF is able to significantly outperform the baseline provided by \textsc{RA} in terms of all AR metrics, on all datasets, models and explanation methods considered in this work. A subset of these results is shown in Figure \ref{fig:stdevalcurves}, the rest can be found in Appendix \ref{apn:robustnessexplanations}. Moreover, we do not find that any architecture is significantly more robust to TEF perturbations for explainers S and IG. However, the self-attention mechanism of transformers seems to be more robust to perturbations than non-transformer-based architectures and explanations. The semantic similarity decreases with increasing $\displaystyle \rho$ and stays above $0.7$ in most cases. This, together with the fact that resulting samples share predictions with the non-perturbed ones effectively highlights that the explanations given by these models and attribution methods lack \textit{faithfulness}.%
\newcommand{\ablationplot}[9]{

    \begin{tikzpicture}
        \begin{axis}[
            width=\plotwidth,
            height=\plotheight,
            xmin=0.01, xmax=0.32,
            ymin=-0.6, ymax=1,
            ymajorgrids=true,
            xmajorgrids=true,
            ytick={-0.5, 0, 0.5 ,1},
            yticklabels={, 0, , 1},
            xtick={0.16, 0.32},
            xticklabels={0.16, $\displaystyle \rho$},
            grid style=dashed,
            legend pos=south west,
            legend columns=2
        ]
    
        \addplot[name path=pcc, r30, mark=*, line width=1pt, opacity=\lineopacity] table[x={x}, y={pcc}] {data/tef/#1/#2/#3.tex};\label{pgf:ablationtef};
        \addlegendentry{TEF}
        \addplot[name path=pccra, r31, mark=triangle, line width=1pt, opacity=\lineopacity] table[x={x}, y={pcc}] {data/ra/#1/#2/#3.tex};\label{pgf:ablationra};
        \addlegendentry{RA}
        \addplot[name path=pccro, b30, mark=o, line width=1pt, opacity=\lineopacity] table[x={x}, y={pcc}] {data/ri/#1/#2/#3.tex};\label{pgf:ablationri};
        \addlegendentry{RI}
        \addplot[name path=pccrs, b31, mark=diamond, line width=1pt, opacity=\lineopacity] table[x={x}, y={pcc}] {data/rs/#1/#2/#3.tex};\label{pgf:ablationrs};
        \addlegendentry{RS}
        
        \addplot[name path=lpcc, r30!50, opacity=\stdopacity] table[x={x}, y={lpcc}] {data/tef/#1/#2/#3.tex};
        \addplot[name path=upcc, r30!50, opacity=\stdopacity] table[x={x}, y={upcc}] {data/tef/#1/#2/#3.tex};
        \addplot[r30!50,fill opacity=\stdopacity] fill between[of=lpcc and upcc];
    
        \addplot[name path=lpcc, r31!50, opacity=\stdopacity] table[x={x}, y={lpcc}] {data/ra/#1/#2/#3.tex};
        \addplot[name path=upcc, r31!50, opacity=\stdopacity] table[x={x}, y={upcc}] {data/ra/#1/#2/#3.tex};
        \addplot[r31!50,fill opacity=\stdopacity] fill between[of=lpcc and upcc];
        
        \addplot[name path=lpcc, b30!50, opacity=\stdopacity] table[x={x}, y={lpcc}] {data/ri/#1/#2/#3.tex};
        \addplot[name path=upcc, b30!50, opacity=\stdopacity] table[x={x}, y={upcc}] {data/ri/#1/#2/#3.tex};
        \addplot[b30!50,fill opacity=\stdopacity] fill between[of=lpcc and upcc];
        
        \addplot[name path=lpcc, b31!50, opacity=\stdopacity] table[x={x}, y={lpcc}] {data/rs/#1/#2/#3.tex};
        \addplot[name path=upcc, b31!50, opacity=\stdopacity] table[x={x}, y={upcc}] {data/rs/#1/#2/#3.tex};
        \addplot[b31!50,fill opacity=\stdopacity] fill between[of=lpcc and upcc];
    
        \node at (axis cs:0.32,#5) [anchor=north east, opacity=\lineopacity] {\scriptsize\textbf{\textcolor{b30}{ACC}: #4}};
        \node at (axis cs:0.32,#7) [anchor=north east, opacity=\lineopacity] {\scriptsize\textbf{\textcolor{b31}{ACC}: #6}};
        \node at (axis cs:0.32,#9) [anchor=north east, opacity=\lineopacity] {\scriptsize\textbf{\textcolor{r31}{ACC}: #8}};
    
        \end{axis}
    \end{tikzpicture}

}

\renewcommand{\ablationlegend}{
	\begin{tikzpicture}
    	\node[draw,fill=white,inner sep=5pt,above left=0.5em] at (0,0) {
    		\begin{tabular}{c}
    			\ref{pgf:ablationtef} TEF	\\
    			\ref{pgf:ablationra} RA		\\
    			\ref{pgf:ablationri} RI		\\
    			\ref{pgf:ablationrs} RS		\\
    		\end{tabular}};
	\end{tikzpicture}
}

\addtolength{\tabcolsep}{-6pt}  

\setlength{\plotwidth}{0.4\textwidth}
\setlength{\plotheight}{0.26\textwidth}

\begin{figure}[t]
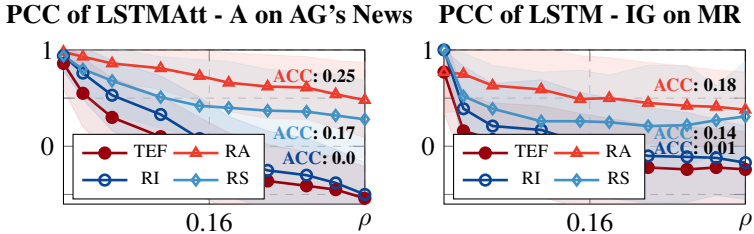

    \centering
    \begin{tabular}{cc}
        \textbf{PCC of LSTMAtt - A on AG's News}		& \textbf{PCC of LSTM - IG on MR}	\\
        \ablationplot{agnews}{lstmatt}{A}{0.0}{0.05}{0.17}{0.3}{0.25}{0.9}
        	& \ablationplot{mr}{lstm}{IG}{0.01}{0.15}{0.14}{0.3}{0.18}{0.8}

    \end{tabular}
    \caption{Ablation study of TEF. We compare the PCC of TEF, RA, the \textsc{RandomImportance} (RI) attack and the \textsc{RandomSynonym} (RS) attack. We find that RI behaves slightly worse than TEF, while RS behaves slightly better than RA in terms of reducing attribution correlation over all $\displaystyle \rho$ values.}
    \label{fig:tefstepinfluence}
\end{figure}

\addtolength{\tabcolsep}{6pt}  
\paragraph{Ablation study.} In addition to the fully random attack described in the previous paragraph, we compare \textsc{TEF} to our semi-random attacks \textsc{RandomImportance} (RI) and \textsc{RandomSynonym} (RS). We randomize the word importance ranking of TEF (RI) but keep the selection of best final synonym, and we randomize the final synonym selection of TEF (RS) but keep the word importance ranking respectively. Figure \ref{fig:tefstepinfluence} shows our findings for these experiments, along with comparisons to \textsc{RandomAttack} (RA). The PCC curves and the ACC values show that RI consistently outperforms RS in terms of PCC over the whole operation interval of $\displaystyle \rho$. Moreover, the impact of word importance ranking diminishes with increasing $\displaystyle \rho$, especially for shorter datasets like MR. This can be observed by RS performing closer to RA for high $\displaystyle \rho$ values.%
\paragraph{BERT's attention layers and heads.} BERT's attention weights can be used to help gain insight into a models prediction by understanding which parts of the input are most \textit{attended} to \citep{bertviz}. Our BERT models have 12 layers with 12 attention heads (144 heads in total), each producing a distribution of attention weights over its inputs and outputs. Estimating the AR of all heads together is not useful, as effects would average out. Therefore, we run TEF to estimate the robustness of each head separately. Figure \ref{fig:bertheadrobustness} contains the average PCCs of the attention weights before and after perturbing the inputs with TEF. We find that attention weights in later layers tend to be more susceptible to input perturbations than earlier layers. Moreover, heads within a layer tend to be comparably robust. We leave a thorough, theoretical analysis of this phenomenon to future work. We conclude that the increasing reliance on attention weights to provide inherent interpretations to BERT predictions needs careful investigation, especially in safety-critical applications.%
%
\newcommand{\attentionheadfigure}[2]{
    \pgfplotstableread{data/attention_heads/#1/#2.tex}{\datamatrix}

    \ifthenelse{\equal{#2}{xlnet}}
    {
        \begin{tikzpicture}[box/.style={rectangle}]
            \begin{axis}[
              width=2\plotwidth,
              height=\plotheight,
              xmin=0.5, xmax=24.5,
              xtick={8, 16, 24},
              xticklabels={8, 16, L.},
              ymin=0.5, ymax=16.5,
              ytick={5, 10, 16},
              yticklabels={5, 10, H.},
            ]
            \foreach \l in {0,...,23}{
                \foreach \h in {0,...,15}{
                    \pgfplotstablegetelem{\l}{\h}\of\datamatrix
                    \edef\temp{\noexpand\node[box,fill=r30!\pgfplotsretval] at (\l+1,\h+1){};}\temp
                };
            };
            \end{axis}
        \end{tikzpicture}
    }
    {
        \begin{tikzpicture}[box/.style={rectangle}]
            \begin{axis}[
              width=\plotwidth,
              height=\plotheight,
              xmin=0.5, xmax=12.5,
              xtick={4, 8, 12},
              xticklabels={4, 8, L.},
              ymin=0.5, ymax=12.5,
              ytick={4, 8, 12},
              yticklabels={4, 8, H.},
            ]
            \foreach \l in {0,...,11}{
                \foreach \h in {0,...,11}{
                    \pgfplotstablegetelem{\l}{\h}\of\datamatrix
                    \edef\temp{\noexpand\node[box,fill=r30!\pgfplotsretval] at (\l+1,\h+1){};}\temp
                };
            };
            \end{axis}
        \end{tikzpicture}
    }%
}

\addtolength{\tabcolsep}{-8pt}  
\setlength{\plotwidth}{0.27\textwidth}
\setlength{\plotheight}{0.27\textwidth}

\begin{figure}[t]
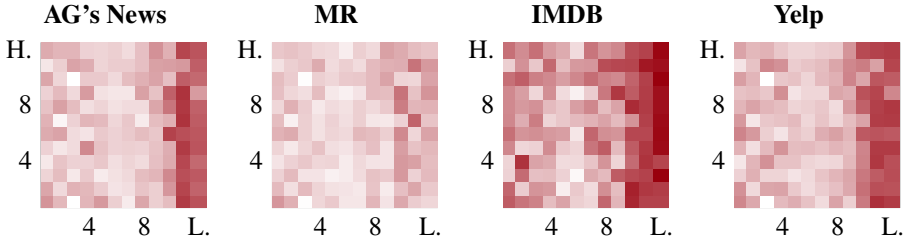

    \centering
    \begin{tabular}{cccc}
    	\textbf{AG's News} & \textbf{MR} & \textbf{IMDB} & \textbf{Yelp} \\
        \attentionheadfigure{agnews}{bert} & \attentionheadfigure{mr}{bert}  & \attentionheadfigure{imdb}{bert} & \attentionheadfigure{yelp}{bert}
    \end{tabular}
    \caption{Estimated robustness of BERT attention weights on different layers (X-axis) and heads (Y-axis) for $\displaystyle \rho_{max} = 0.2$. Red cells indicate average PCC values close to 0, hence less robust attention head weights, while white cells have average PCCs close to 1. Attention heads in later layers tend to be less robust, while heads within a layer seem equally robust in most layers.}
    \label{fig:bertheadrobustness}
\end{figure}

\addtolength{\tabcolsep}{8pt}  
\subsection{Transferability and semi-universal perturbations}%
\addtolength{\tabcolsep}{-8pt}  

\setlength{\plotwidth}{0.4\textwidth}
\setlength{\plotheight}{0.28\textwidth}

\begin{figure}[t]
    \centering
    \footnotesize
    	\begin{tabular}{ccc}
    		\textbf{PCC of LSTMAtt - IG on IMDB}
    		& \textbf{PCC of CNN - IG on AG's News}
    		& \textbf{PCC of CNN - IG on Yelp} \\
    		\begin{tikzpicture}
        \begin{axis}[
            width=\plotwidth,
            height=\plotheight,
            xmin=0.01, xmax=0.32,
            ymin=-0.6, ymax=1,
            ymajorgrids=true,
            xmajorgrids=true,
            ytick={-0.5, 0, 0.5 ,1},
            yticklabels={, 0, , 1},
            xtick={0.16, 0.32},
            xticklabels={0.16, $\displaystyle \rho$},
            grid style=dashed,
            legend pos=south west
            ]

        \addplot[name path=lstmig, b30, line width=1pt, mark=o, opacity=\lineopacity] table[x={x}, y={pcc}] {data/transfer/imdb/lstmatt/IG/lstm_IG.tex};
        \addlegendentry{LSTM-IG}
        \addplot[name path=lstmatts, b31, line width=1pt, mark=diamond, opacity=\lineopacity] table[x={x}, y={pcc}] {data/transfer/imdb/lstmatt/IG/lstmatt_S.tex};
        \addlegendentry{LSTMAtt-S}
        \addplot[name path=lstmatta, b32, line width=1pt, mark=pentagon, opacity=\lineopacity] table[x={x}, y={pcc}] {data/transfer/imdb/lstmatt/IG/lstmatt_A.tex};
        \addlegendentry{LSTMAtt-A}
        
        \addplot[name path=tef, r30, line width=1pt, mark=*, opacity=\lineopacity] table[x={x}, y={pcc}] {data/tef/imdb/lstmatt/IG.tex};
        \addplot[name path=ra, r31, line width=1pt, mark=triangle, opacity=\lineopacity] table[x={x}, y={pcc}] {data/ra/imdb/lstmatt/IG.tex};
        
        \addplot[name path=ltef, r30!50, opacity=\stdopacity] table[x={x}, y={lpcc}] {data/tef/imdb/lstmatt/IG.tex};
        \addplot[name path=utef, r30!50, opacity=\stdopacity] table[x={x}, y={upcc}] {data/tef/imdb/lstmatt/IG.tex};
        \addplot[r30!50,fill opacity=\stdopacity] fill between[of=ltef and utef];
        
        \addplot[name path=lra, r31!50, opacity=\stdopacity] table[x={x}, y={lpcc}] {data/ra/imdb/lstmatt/IG.tex};
        \addplot[name path=ura, r31!50, opacity=\stdopacity] table[x={x}, y={upcc}] {data/ra/imdb/lstmatt/IG.tex};
        \addplot[r31!50,fill opacity=\stdopacity] fill between[of=lra and ura];
        
        
        \addplot[name path=llstmig, b30!50, opacity=\stdopacity] table[x={x}, y={lpcc}] {data/transfer/imdb/lstmatt/IG/lstm_IG.tex};
        \addplot[name path=ulstmig, b30!50, opacity=\stdopacity] table[x={x}, y={upcc}] {data/transfer/imdb/lstmatt/IG/lstm_IG.tex};
        \addplot[b30!50,fill opacity=\stdopacity] fill between[of=llstmig and ulstmig];
        
        \addplot[name path=llstmatta, b31!50, opacity=\stdopacity] table[x={x}, y={lpcc}] {data/transfer/imdb/lstmatt/IG/lstmatt_S.tex};
        \addplot[name path=ulstmatta, b31!50, opacity=\stdopacity] table[x={x}, y={upcc}] {data/transfer/imdb/lstmatt/IG/lstmatt_S.tex};
        \addplot[b31!50,fill opacity=\stdopacity] fill between[of=llstmatta and ulstmatta];
        
        \addplot[name path=llstmatta, b32!50, opacity=\stdopacity] table[x={x}, y={lpcc}] {data/transfer/imdb/lstmatt/IG/lstmatt_A.tex};
        \addplot[name path=ulstmatta, b32!50, opacity=\stdopacity] table[x={x}, y={upcc}] {data/transfer/imdb/lstmatt/IG/lstmatt_A.tex};
        \addplot[b32!50,fill opacity=\stdopacity] fill between[of=llstmatta and ulstmatta];
    
        \node at (axis cs:0.32,-0.38) [anchor=north east, opacity=\lineopacity] {\scriptsize\textbf{\textcolor{r30}{ACC}: -0.07}};
        \node at (axis cs:0.32,0.75) [anchor=north east, opacity=\lineopacity] {\scriptsize\textbf{\textcolor{r31}{ACC}: 0.19}};
        \node at (axis cs:0.32,0.2) [anchor=north east, opacity=\lineopacity] {\scriptsize\textbf{\textcolor{b30}{ACC}: 0.10}};
    
        \end{axis}
    \end{tikzpicture}
    		& \begin{tikzpicture}
        \begin{axis}[
            width=\plotwidth,
            height=\plotheight,
            xmin=0.01, xmax=0.32,
            ymin=-0.6, ymax=1,
            ymajorgrids=true,
            xmajorgrids=true,
            ytick={-0.5, 0, 0.5 ,1},
            yticklabels={, 0, , 1},
            xtick={0.16, 0.32},
            xticklabels={0.16, $\displaystyle \rho$},
            grid style=dashed,
            legend pos=south west
            ]
    
        \addplot[name path=berta, b30, line width=1pt, mark=o, opacity=\lineopacity] table[x={x}, y={pcc}] {data/transfer/agnews/cnn/IG/bert_A.tex};
        \addlegendentry{BERT-A}
        \addplot[name path=lstmig, b31, line width=1pt, mark=diamond, opacity=\lineopacity] table[x={x}, y={pcc}] {data/transfer/agnews/cnn/IG/lstm_IG.tex};
        \addlegendentry{LSTM-IG}
        \addplot[name path=cnns, b32, line width=1pt, mark=pentagon, opacity=\lineopacity] table[x={x}, y={pcc}] {data/transfer/agnews/cnn/IG/cnn_S.tex};
        \addlegendentry{CNN-S}
        
        \addplot[name path=tef, r30, line width=1pt, mark=*, opacity=\lineopacity] table[x={x}, y={pcc}] {data/tef/agnews/cnn/IG.tex};\label{plt:transfertef};
        \addplot[name path=ra, r31, line width=1pt, mark=triangle, opacity=\lineopacity] table[x={x}, y={pcc}] {data/ra/agnews/cnn/IG.tex};\label{plt:transferra};
        
        \addplot[name path=ltef, r30!50, opacity=\stdopacity] table[x={x}, y={lpcc}] {data/tef/agnews/cnn/IG.tex};
        \addplot[name path=utef, r30!50, opacity=\stdopacity] table[x={x}, y={upcc}] {data/tef/agnews/cnn/IG.tex};
        \addplot[r30!50,fill opacity=\stdopacity] fill between[of=ltef and utef];
%
        \addplot[name path=lra, r31!50, opacity=\stdopacity] table[x={x}, y={lpcc}] {data/ra/agnews/cnn/IG.tex};
        \addplot[name path=ura, r31!50, opacity=\stdopacity] table[x={x}, y={upcc}] {data/ra/agnews/cnn/IG.tex};
        \addplot[r31!50,fill opacity=\stdopacity] fill between[of=lra and ura];
        \addplot[name path=llstmig, b30!50, opacity=\stdopacity] table[x={x}, y={lpcc}] {data/transfer/agnews/cnn/IG/bert_A.tex};
        \addplot[name path=ulstmig, b30!50, opacity=\stdopacity] table[x={x}, y={upcc}] {data/transfer/agnews/cnn/IG/bert_A.tex};
        \addplot[b30!50,fill opacity=\stdopacity] fill between[of=llstmig and ulstmig];
        \addplot[name path=llstmig, b31!50, opacity=\stdopacity] table[x={x}, y={lpcc}] {data/transfer/agnews/cnn/IG/lstm_IG.tex};
        \addplot[name path=ulstmig, b31!50, opacity=\stdopacity] table[x={x}, y={upcc}] {data/transfer/agnews/cnn/IG/lstm_IG.tex};
        \addplot[b31!50,fill opacity=\stdopacity] fill between[of=llstmig and ulstmig];
        \addplot[name path=lcnna, b32!50, opacity=\stdopacity] table[x={x}, y={lpcc}] {data/transfer/agnews/cnn/IG/cnn_S.tex};
        \addplot[name path=ucnna, b32!50, opacity=\stdopacity] table[x={x}, y={upcc}] {data/transfer/agnews/cnn/IG/cnn_S.tex};
        \addplot[b32!50,fill opacity=\stdopacity] fill between[of=lcnna and ucnna];

        \node at (axis cs:0.32,-0.25) [anchor=north east, opacity=\lineopacity] {\scriptsize\textbf{\textcolor{r30}{ACC}: 0.06}};
        \node at (axis cs:0.32,0.9) [anchor=north east, opacity=\lineopacity] {\scriptsize\textbf{\textcolor{r31}{ACC}: 0.27}};
        \node at (axis cs:0.32,0.45) [anchor=north east, opacity=\lineopacity] {\scriptsize\textbf{\textcolor{b32}{ACC}: 0.22}};
        \end{axis}
    \end{tikzpicture}
    		& \begin{tikzpicture}
        \begin{axis}[
            width=\plotwidth,
            height=\plotheight,
            xmin=0.01, xmax=0.32,
            ymin=-0.6, ymax=1,
            ymajorgrids=true,
            xmajorgrids=true,
            ytick={-0.5, 0, 0.5 ,1},
            yticklabels={, 0, , 1},
            xtick={0.16, 0.32},
            xticklabels={0.16, $\displaystyle \rho$},
            grid style=dashed,
            legend pos=south west
            ]
    
        \addplot[name path=berta, b30, line width=1pt, mark=o, opacity=\lineopacity] table[x={x}, y={pcc}] {data/transfer/yelp/cnn/IG/bert_A.tex};
        \addlegendentry{BERT-A}
        \addplot[name path=cnns, b31, line width=1pt, mark=diamond, opacity=\lineopacity] table[x={x}, y={pcc}] {data/transfer/yelp/cnn/IG/cnn_S.tex};
        \addlegendentry{CNN-S}
        \addplot[name path=lstmattig, b32, line width=1pt, mark=pentagon, opacity=\lineopacity] table[x={x}, y={pcc}] {data/transfer/yelp/cnn/IG/lstmatt_IG.tex};
        \addlegendentry{LSTMAtt-IG}
        
        \addplot[name path=tef, r30, line width=1pt, mark=*, opacity=\lineopacity] table[x={x}, y={pcc}] {data/tef/yelp/cnn/IG.tex};
        \addplot[name path=ra, r31, line width=1pt, mark=triangle, opacity=\lineopacity] table[x={x}, y={pcc}] {data/ra/yelp/cnn/IG.tex};

        \addplot[name path=ltef, r30!50, opacity=\stdopacity] table[x={x}, y={lpcc}] {data/tef/yelp/cnn/IG.tex};
        \addplot[name path=utef, r30!50, opacity=\stdopacity] table[x={x}, y={upcc}] {data/tef/yelp/cnn/IG.tex};
        \addplot[r30!50,fill opacity=\stdopacity] fill between[of=ltef and utef];
        \addplot[name path=lra, r31!50, opacity=\stdopacity] table[x={x}, y={lpcc}] {data/ra/yelp/cnn/IG.tex};
        \addplot[name path=ura, r31!50, opacity=\stdopacity] table[x={x}, y={upcc}] {data/ra/yelp/cnn/IG.tex};
        \addplot[r31!50,fill opacity=\stdopacity] fill between[of=lra and ura];
        \addplot[name path=lcnnig, b30!50, opacity=\stdopacity] table[x={x}, y={lpcc}] {data/transfer/yelp/cnn/IG/bert_A.tex};
        \addplot[name path=ucnnig, b30!50, opacity=\stdopacity] table[x={x}, y={upcc}] {data/transfer/yelp/cnn/IG/bert_A.tex};
        \addplot[b30!50,fill opacity=\stdopacity] fill between[of=lcnnig and ucnnig];
        \addplot[name path=lcnna, b31!50, opacity=\stdopacity] table[x={x}, y={lpcc}] {data/transfer/yelp/cnn/IG/cnn_S.tex};
        \addplot[name path=ucnna, b31!50, opacity=\stdopacity] table[x={x}, y={upcc}] {data/transfer/yelp/cnn/IG/cnn_S.tex};
        \addplot[b31!50,fill opacity=\stdopacity] fill between[of=lcnna and ucnna];
        \addplot[name path=lcnna, b32!50, opacity=\stdopacity] table[x={x}, y={lpcc}] {data/transfer/yelp/cnn/IG/lstmatt_IG.tex};
        \addplot[name path=ucnna, b32!50, opacity=\stdopacity] table[x={x}, y={upcc}] {data/transfer/yelp/cnn/IG/lstmatt_IG.tex};
        \addplot[b32!50,fill opacity=\stdopacity] fill between[of=lcnna and ucnna];
    
        \node at (axis cs:0.32,-0.2) [anchor=north east, opacity=\lineopacity, color=black] {\scriptsize\textbf{\textcolor{r30}{ACC}: -0.09}};
        \node at (axis cs:0.32,0.6) [anchor=north east, opacity=\lineopacity, color=black] {\scriptsize\textbf{\textcolor{r31}{ACC}: 0.15}};
        \node at (axis cs:0.32,0.15) [anchor=north east, opacity=\lineopacity, color=black] {\scriptsize\textbf{\textcolor{b31}{ACC}: 0.09}};
    
        \end{axis}
    \end{tikzpicture}\\
    	\end{tabular}
    
    \caption{Transfer capabilities of TEF to other models and explanation methods. The lines indicate the estimated PCC of TEF perturbations transferred from the indicated models and explanations. \ref{plt:transfertef} and \ref{plt:transferra} indicate the PCC curve of optimal TEF and RA perturbations respectively, without transfer.}
    \label{fig:transferabilityfig}
\end{figure}
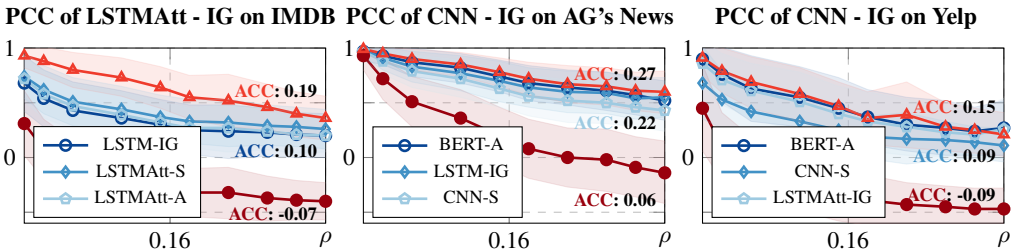

\addtolength{\tabcolsep}{8pt}  
%
%
\paragraph{Transferability of perturbations to models and explanations.} The adversary does not necessarily possess information about the deployec model nor the exact method to produce the accompanying explanations. Therefore, it is crucial for systems to be as resistant to transfer attacks as possible in order to evade perturbations constructed on similar models and explanations.\\%
Thus, we examine how our classifiers and attribution methods react to transfer attacks computed with TEF. We alter the input samples for a given model and explanation method with TEF, then evaluate the PCC of attributions on the same samples but different architectures and explainers. The results are found in Figure \ref{fig:transferabilityfig}. We observe that transfer attacks perform better than RA, some even by approx. $0.4$ in terms of average PCC decrease in the operation area of $\displaystyle \rho \approx 0.1$. However, as expected, they significantly fall short of the performance of TEF. Therefore, we conclude that transferring TEF perturbations across models and explainers effectively highlights fragility of explanations, but TEF provides tighter AR bounds without transfer.%
\paragraph{Semi-universal perturbations.} In this section, we take a step towards defining universal perturbations, similarly to the work of the authors  \cite{universalimage} and \cite{universaltext}. These provide fast and computationally cheap perturbations during attack time that are able to mislead classifiers with pre-computed perturbations. However, we attack the explanations of text classifiers, instead of their predictions and call our perturbations \textit{semi-universal attack policies}.\\%
We split the test dataset into two equally sized parts, the attack set and the evaluation set. We utilize the former for constructing our \textit{semi-universal attack policies} and the latter to evaluate how effectively our semi-universal attack alters the attributions maps of our models.\\%
First, for each sample in the attack set, we compute the optimal TEF perturbation for all our models and explainers. We then extract statistics of these perturbations, which are the most common replacement and the replacement frequency for each replaced token and sort these by decreasing frequency. These are our \textit{semi-universal attack policies}, seen in Figure \ref{fig:attackpolicies}. During this \textit{construction} phase, we query the model for predictions and explanations, as we compute optimal TEF perturbations.\\%
Second, we \textit{evaluate} our semi-universal attack that utilizes the aforementioned policies to alter explanations of classifiers. The inputs to this attack are a text sample, a semi-universal policy and a maximum perturbed ratio $\displaystyle \rho_{max}$. The attack iterates over the policy, starting with the token in the first row and finishing with the last. Whenever the current token is found in the input text sample, it is replaced with the replacement token in the list. If the perturbed ratio exceeds $\displaystyle \rho_{max}$, the attack is aborted. In such a way, perturbed inputs are created without querying the model during attack time. The actual perturbation for each text sample depends on the sample, hence the name \textit{semi-universal attack policy}. The resulting samples are then evaluated on a given model and explanation method. Representative results are given in Figure \ref{fig:universalcurves}. We conclude that our semi-universal policies are effective in reducing attribution correlation when the adversary has no access to the target model and explanation method, as indicated by the lower ACC values of the semi-universal PCC curves.
\addtolength{\tabcolsep}{-5pt}  

\setlength{\plotwidth}{0.4\textwidth}
\setlength{\plotheight}{0.3\textwidth}

\begin{figure}[t]
    \centering
    \footnotesize
    \begin{tabular}{ccc}
    	\textbf{AG's News}
    	& \textbf{IMDB}
    	& \textbf{MR}	\\
    	\begin{tabular}{c|c|c}
\textbf{Token}       & \textbf{Repl. \#} & \textbf{Replacement} \\\hline
reuters     & 146k  & goldman       \\
said        & 131k  & avowed        \\
new         & 130k  & nouvelle \\
ap          & 107k  & ha\\
oil          & 72.8k  & tar\\
...         & ...   & ...                   \\
workers   & 10.9k & labourers  \\
...         & ...   & ...                   \\
zone        & 2.9k  & field
\end{tabular}
    	& \begin{tabular}{c|c|c}
\textbf{Token}       & \textbf{Repl. \#} & \textbf{Replacement} \\\hline
movie     & 430k  & cinematographic       \\
film        & 338k  & cine        \\
good         & 122k  & decent \\
great          & 103k  & whopping\\
bad          & 102k  & wicked\\
...         & ...   & ...                   \\
amazing   & 17.1k & staggering  \\
...         & ...   & ...                   \\
scary        & 6.8k  & fearful
\end{tabular}
    	& \begin{tabular}{c|c|c}
\textbf{Token}       & \textbf{Repl. \#} & \textbf{Replacement} \\\hline
movie     & 8.2k  & cinematographic       \\
film        & 8.0k  & cinematographic        \\
story         & 2.6k  & conte \\
good          & 2.5k  & decent\\
comedy          & 2.4k  & humorist\\
...         & ...   & ...                   \\
triumph   & 139 & victory  \\
...         & ...   & ...                   \\
shines        & 69  & glows
\end{tabular}
    \end{tabular}
    \caption{Semi-universal attack policies for different datasets.}
    \label{fig:attackpolicies}
\end{figure}

\addtolength{\tabcolsep}{5pt}  
\addtolength{\tabcolsep}{-5pt}  

\setlength{\plotwidth}{0.4\textwidth}
\setlength{\plotheight}{0.3\textwidth}

\begin{figure}[t]
    \centering
    \footnotesize
    \begin{tabular}{ccc}
    	\textbf{PCC of LSTM - IG on AG's News}
    	& \textbf{PCC of LSTM - IG on IMDB}
    	& \textbf{PCC of LSTMAtt - IG on MR}	\\
    	\begin{tikzpicture}
    \begin{axis}[
        width=\plotwidth,
        height=\plotheight,
        xmin=0.01, xmax=0.32,
        ymin=-0.6, ymax=1,
        ymajorgrids=true,
        xmajorgrids=true,
        ytick={-0.5, 0, 0.5 ,1},
        yticklabels={, 0, , 1},
        xtick={0.16, 0.32},
        xticklabels={0.16, $\displaystyle \rho$},
        grid style=dashed,
        legend pos=south west,
        legend columns=1
        ]

    \addplot[name path=tef, r30, line width=1pt, mark=*, opacity=\lineopacity] table[x={x}, y={pcc}] {data/tef/agnews/lstm/IG.tex};
    \addlegendentry{TEF}
    \addplot[name path=ra, r31, line width=1pt, mark=triangle, opacity=\lineopacity] table[x={x}, y={pcc}] {data/ra/agnews/lstm/IG.tex};
    \addlegendentry{RA}
    \addplot[name path=uni, b30, line width=1pt, mark=o, opacity=\lineopacity] table[x={x}, y={pcc}] {data/universal/agnews/lstm/IG_uni.tex};
    \addlegendentry{Semi-uni.}
    
    \addplot[name path=ltef, r30!50, opacity=\stdopacity] table[x={x}, y={lpcc}] {data/tef/agnews/lstm/IG.tex};
    \addplot[name path=utef, r30!50, opacity=\stdopacity] table[x={x}, y={upcc}] {data/tef/agnews/lstm/IG.tex};
    \addplot[r30!50,fill opacity=\stdopacity] fill between[of=ltef and utef];
    
    \addplot[name path=lra, r31!50, opacity=\stdopacity] table[x={x}, y={lpcc}] {data/ra/agnews/lstm/IG.tex};
    \addplot[name path=ura, r31!50, opacity=\stdopacity] table[x={x}, y={upcc}] {data/ra/agnews/lstm/IG.tex};
    \addplot[r31!50,fill opacity=\stdopacity] fill between[of=lra and ura];
    
    
    \addplot[name path=luni, b30!50, opacity=\stdopacity] table[x={x}, y={lpcc}] {data/universal/agnews/lstm/IG_uni.tex};
    \addplot[name path=uuni, b30!50, opacity=\stdopacity] table[x={x}, y={upcc}] {data/universal/agnews/lstm/IG_uni.tex};
    \addplot[b30!50,fill opacity=\stdopacity] fill between[of=luni and uuni];

    \node at (axis cs:0.32,-0.27) [anchor=north east, opacity=\lineopacity] {\scriptsize\textbf{\textcolor{r30}{ACC}: 0.00}};
    \node at (axis cs:0.32,0.9) [anchor=north east, opacity=\lineopacity] {\scriptsize\textbf{\textcolor{r31}{ACC}: 0.26}};
    \node at (axis cs:0.32,0.4) [anchor=north east, opacity=\lineopacity] {\scriptsize\textbf{\textcolor{b30}{ACC}: 0.20}};

    \end{axis}
\end{tikzpicture}
    	& \begin{tikzpicture}
    \begin{axis}[
        width=\plotwidth,
        height=\plotheight,
        xmin=0.01, xmax=0.32,
        ymin=-0.6, ymax=1,
        ymajorgrids=true,
        xmajorgrids=true,
        ytick={-0.5, 0, 0.5 ,1},
        yticklabels={, 0, , 1},
        xtick={0.16, 0.32},
        xticklabels={0.16, $\displaystyle \rho$},
        grid style=dashed,
        legend pos=south west,
        legend columns=1
        ]

    \addplot[name path=tef, r30, line width=1pt, mark=*, opacity=\lineopacity] table[x={x}, y={pcc}] {figures/universal/imdb/data/lstm_IG/tef.dat};
    \addlegendentry{TEF}
    \addplot[name path=ra, r31, line width=1pt, mark=triangle, opacity=\lineopacity] table[x={x}, y={pcc}] {figures/universal/imdb/data/lstm_IG/ra.dat};
    \addlegendentry{RA}
    \addplot[name path=uni, b30, line width=1pt, mark=o, opacity=\lineopacity] table[x={x}, y={pcc}] {figures/universal/imdb/data/lstm_IG/uni.dat};
    \addlegendentry{Semi-uni.}
    
    \addplot[name path=ltef, r30!50, opacity=\stdopacity] table[x={x}, y={lpcc}] {figures/universal/imdb/data/lstm_IG/tef.dat};
    \addplot[name path=utef, r30!50, opacity=\stdopacity] table[x={x}, y={upcc}] {figures/universal/imdb/data/lstm_IG/tef.dat};
    \addplot[r30!50,fill opacity=\stdopacity] fill between[of=ltef and utef];
    
    \addplot[name path=lra, r31!50, opacity=\stdopacity] table[x={x}, y={lpcc}] {figures/universal/imdb/data/lstm_IG/ra.dat};
    \addplot[name path=ura, r31!50, opacity=\stdopacity] table[x={x}, y={upcc}] {figures/universal/imdb/data/lstm_IG/ra.dat};
    \addplot[r31!50,fill opacity=\stdopacity] fill between[of=lra and ura];
    
    
    \addplot[name path=luni, b30!50, opacity=\stdopacity] table[x={x}, y={lpcc}] {figures/universal/imdb/data/lstm_IG/uni.dat};
    \addplot[name path=uuni, b30!50, opacity=\stdopacity] table[x={x}, y={upcc}] {figures/universal/imdb/data/lstm_IG/uni.dat};
    \addplot[b30!50,fill opacity=\stdopacity] fill between[of=luni and uuni];
    
    \node at (axis cs:0.32,-0.27) [anchor=north east, opacity=\lineopacity, color=black] {\scriptsize\textbf{\textcolor{r30}{ACC}: -0.04}};
    \node at (axis cs:0.32,0.85) [anchor=north east, opacity=\lineopacity, color=black] {\scriptsize\textbf{\textcolor{r31}{ACC}: 0.22}};
    \node at (axis cs:0.32,0.13) [anchor=north east, opacity=\lineopacity, color=black] {\scriptsize\textbf{\textcolor{b30}{ACC}: 0.14}};

    \end{axis}
\end{tikzpicture}
    	& \begin{tikzpicture}
    \begin{axis}[
        width=\plotwidth,
        height=\plotheight,
        xmin=0.01, xmax=0.32,
        ymin=-0.6, ymax=1,
        ymajorgrids=true,
        xmajorgrids=true,
        ytick={-0.5, 0, 0.5 ,1},
        yticklabels={, 0, , 1},
        xtick={0.16, 0.32},
        xticklabels={0.16, $\displaystyle \rho$},
        grid style=dashed,
        legend pos=south west
        ]

    \addplot[name path=tef, r30, line width=1pt, mark=*, opacity=\lineopacity] table[x={x}, y={pcc}] {data/tef/mr/lstmatt/IG.tex};
    \addlegendentry{TEF}
    \addplot[name path=ra, r31, line width=1pt, mark=triangle, opacity=\lineopacity] table[x={x}, y={pcc}] {data/ra/mr/lstmatt/IG.tex};
    \addlegendentry{RA}
    \addplot[name path=uni, b30, line width=1pt, mark=o, opacity=\lineopacity] table[x={x}, y={pcc}] {data/universal/mr/lstmatt/IG_uni.tex};
    \addlegendentry{Semi-uni.}
    
	\addplot[name path=ltef, r30!50, opacity=\stdopacity] table[x={x}, y={lpcc}] {data/tef/mr/lstmatt/IG.tex};
    \addplot[name path=utef, r30!50, opacity=\stdopacity] table[x={x}, y={upcc}] {data/tef/mr/lstmatt/IG.tex};
    \addplot[r30!50,fill opacity=\stdopacity] fill between[of=ltef and utef];
    
    \addplot[name path=lra, r31!50, opacity=\stdopacity] table[x={x}, y={lpcc}] {data/ra/mr/lstmatt/IG.tex};
    \addplot[name path=ura, r31!50, opacity=\stdopacity] table[x={x}, y={upcc}] {data/ra/mr/lstmatt/IG.tex};
    \addplot[r31!50,fill opacity=\stdopacity] fill between[of=lra and ura];
    
    
    \addplot[name path=luni, b30!50, opacity=\stdopacity] table[x={x}, y={lpcc}] {data/universal/mr/lstmatt/IG_uni.tex};
    \addplot[name path=uuni, b30!50, opacity=\stdopacity] table[x={x}, y={upcc}] {data/universal/mr/lstmatt/IG_uni.tex};
    \addplot[b30!50,fill opacity=\stdopacity] fill between[of=luni and uuni];

    \node at (axis cs:0.32,-0.27) [anchor=north east, opacity=\lineopacity, color=black] {\scriptsize\textbf{\textcolor{r30}{ACC}: 0.05}};
    \node at (axis cs:0.32,0.9) [anchor=north east, opacity=\lineopacity, color=black] {\scriptsize\textbf{\textcolor{r31}{ACC}: 0.26}};
    \node at (axis cs:0.32,0.4) [anchor=north east, opacity=\lineopacity, color=black] {\scriptsize\textbf{\textcolor{b30}{ACC}: 0.20}};

    \end{axis}
\end{tikzpicture}
    \end{tabular}
    \caption{Average PCC of the indicated architectures and explainers after applying the semi-universal perturbations (Semi-uni.), compared to TEF and RA attacks. The semi-universal attack successfully decreases the correlation of original and attacked attribution maps.}
    \label{fig:universalcurves}
\end{figure}
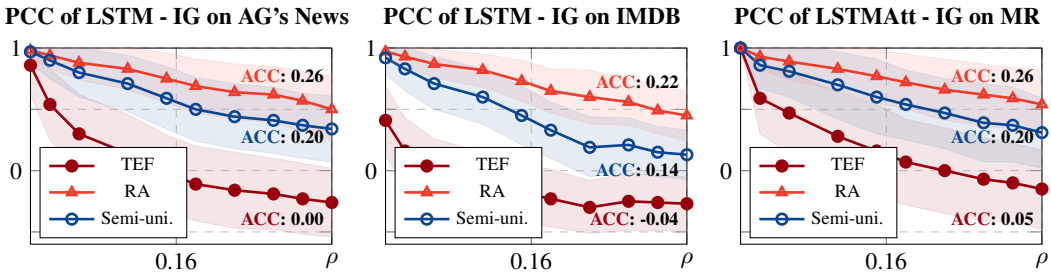

\addtolength{\tabcolsep}{5pt}  

\section{Conclusion}
\label{sec:conclusion}
In this work, we introduced a novel black-box attack called \textsc{TextExplanationFooler}, that successfully perturbs input data such that the outcome of popular explanation methods in sequence classification, but not the prediction of the classifier. This attack provides a baseline estimator for attribution robustness and highlights the lack of robustness of current explanation methods. We compared it to the random attack, showing its superior performance to it on five different, widely used text classification datasets. Moreover, the transfer capabilities of the attack are evaluated. Finally, we showed the existence of semi-universal perturbation policies that are capable of altering explanations without querying the model during attack-time, even without having access to perturbations for those models. In future work, we plan to examine whether a similar white-box attack that has access to model gradients can improve robustness estimation. Moreover, instead of synonym embeddings, we plan to use BERT-based masked language models to extract possible candidates, further improving imperceptible word substitutions.

\bibliography{iclr2022_conference}

\begin{thebibliography}{58}
\providecommand{\natexlab}[1]{#1}
\providecommand{\url}[1]{\texttt{#1}}
\expandafter\ifx\csname urlstyle\endcsname\relax
  \providecommand{\doi}[1]{doi: #1}\else
  \providecommand{\doi}{doi: \begingroup \urlstyle{rm}\Url}\fi

\bibitem[Adadi \& Berrada(2020)Adadi and Berrada]{aihealthcare}
Amina Adadi and Mohammed Berrada.
\newblock Explainable ai for healthcare: from black box to interpretable
  models.
\newblock In \emph{Embedded Systems and Artificial Intelligence}, pp.\
  327--337. Springer, 2020.

\bibitem[Asghar(2016)]{yelp}
Nabiha Asghar.
\newblock Yelp dataset challenge: Review rating prediction.
\newblock \emph{arXiv preprint arXiv:1605.05362}, 2016.

\bibitem[Bahdanau et~al.(2014)Bahdanau, Cho, and Bengio]{attentionmechanism}
Dzmitry Bahdanau, Kyunghyun Cho, and Yoshua Bengio.
\newblock Neural machine translation by jointly learning to align and
  translate.
\newblock \emph{arXiv preprint arXiv:1409.0473}, 2014.

\bibitem[Binder et~al.(2021)Binder, Bockmayr, H{\"a}gele, Wienert, Heim,
  Hellweg, Ishii, Stenzinger, Hocke, Denkert, et~al.]{precisionmedicine}
Alexander Binder, Michael Bockmayr, Miriam H{\"a}gele, Stephan Wienert, Daniel
  Heim, Katharina Hellweg, Masaru Ishii, Albrecht Stenzinger, Andreas Hocke,
  Carsten Denkert, et~al.
\newblock Morphological and molecular breast cancer profiling through
  explainable machine learning.
\newblock \emph{Nature Machine Intelligence}, 3\penalty0 (4):\penalty0
  355--366, 2021.

\bibitem[Buckman et~al.(2018)Buckman, Roy, Raffel, and
  Goodfellow]{thermometerencoding}
Jacob Buckman, Aurko Roy, Colin Raffel, and Ian Goodfellow.
\newblock Thermometer encoding: One hot way to resist adversarial examples.
\newblock In \emph{International Conference on Learning Representations}, 2018.

\bibitem[Carlini \& Wagner(2017)Carlini and Wagner]{carliniwagnerattack}
Nicholas Carlini and David Wagner.
\newblock Towards evaluating the robustness of neural networks.
\newblock In \emph{2017 ieee symposium on security and privacy (sp)}, pp.\
  39--57. IEEE, 2017.

\bibitem[Cer et~al.(2018)Cer, Yang, Kong, Hua, Limtiaco, John, Constant,
  Guajardo-C{\'e}spedes, Yuan, Tar, et~al.]{use}
Daniel Cer, Yinfei Yang, Sheng-yi Kong, Nan Hua, Nicole Limtiaco, Rhomni~St
  John, Noah Constant, Mario Guajardo-C{\'e}spedes, Steve Yuan, Chris Tar,
  et~al.
\newblock Universal sentence encoder.
\newblock \emph{arXiv preprint arXiv:1803.11175}, 2018.

\bibitem[Chen et~al.(2019)Chen, Wu, Rastogi, Liang, and Jha]{igsum}
Jiefeng Chen, Xi~Wu, Vaibhav Rastogi, Yingyu Liang, and Somesh Jha.
\newblock {Robust Attribution Regularization}.
\newblock In \emph{{Advances in Neural Information Processing Systems}}, pp.\
  14300--14310, 2019.

\bibitem[Choi et~al.(2016)Choi, Bahadori, Kulas, Schuetz, Stewart, and
  Sun]{attentionhealthcare}
Edward Choi, Mohammad~Taha Bahadori, Joshua~A Kulas, Andy Schuetz, Walter~F
  Stewart, and Jimeng Sun.
\newblock Retain: An interpretable predictive model for healthcare using
  reverse time attention mechanism.
\newblock \emph{arXiv preprint arXiv:1608.05745}, 2016.

\bibitem[Cisse et~al.(2017)Cisse, Bojanowski, Grave, Dauphin, and
  Usunier]{parsevalnetworks}
Moustapha Cisse, Piotr Bojanowski, Edouard Grave, Yann Dauphin, and Nicolas
  Usunier.
\newblock Parseval networks: Improving robustness to adversarial examples.
\newblock In \emph{International Conference on Machine Learning}, pp.\
  854--863. PMLR, 2017.

\bibitem[Dombrowski et~al.(2019)Dombrowski, Alber, Anders, Ackermann,
  M{\"u}ller, and Kessel]{geometryblame}
Ann-Kathrin Dombrowski, Maximillian Alber, Christopher Anders, Marcel
  Ackermann, Klaus-Robert M{\"u}ller, and Pan Kessel.
\newblock Explanations can be manipulated and geometry is to blame.
\newblock In \emph{Advances in Neural Information Processing Systems}, pp.\
  13589--13600, 2019.

\bibitem[Ebrahimi et~al.(2017)Ebrahimi, Rao, Lowd, and Dou]{hotflip}
Javid Ebrahimi, Anyi Rao, Daniel Lowd, and Dejing Dou.
\newblock Hotflip: White-box adversarial examples for text classification.
\newblock \emph{arXiv preprint arXiv:1712.06751}, 2017.

\bibitem[Etmann et~al.(2019)Etmann, Lunz, Maass, and
  Sch{\"o}nlieb]{alignmentconnection}
Christian Etmann, Sebastian Lunz, Peter Maass, and Carola-Bibiane
  Sch{\"o}nlieb.
\newblock On the connection between adversarial robustness and saliency map
  interpretability.
\newblock \emph{arXiv preprint arXiv:1905.04172}, 2019.

\bibitem[Feng et~al.(2018)Feng, Wallace, Grissom~II, Iyyer, Rodriguez, and
  Boyd-Graber]{nnpathologies}
Shi Feng, Eric Wallace, Alvin Grissom~II, Mohit Iyyer, Pedro Rodriguez, and
  Jordan Boyd-Graber.
\newblock Pathologies of neural models make interpretations difficult.
\newblock \emph{arXiv preprint arXiv:1804.07781}, 2018.

\bibitem[Gao \& Oates(2019)Gao and Oates]{universaltext}
Hang Gao and Tim Oates.
\newblock Universal adversarial perturbation for text classification.
\newblock \emph{arXiv preprint arXiv:1910.04618}, 2019.

\bibitem[Ghaeini et~al.(2018)Ghaeini, Fern, and Tadepalli]{attentioninherent}
Reza Ghaeini, Xiaoli~Z Fern, and Prasad Tadepalli.
\newblock Interpreting recurrent and attention-based neural models: a case
  study on natural language inference.
\newblock \emph{arXiv preprint arXiv:1808.03894}, 2018.

\bibitem[Ghorbani et~al.(2019)Ghorbani, Abid, and Zou]{interpretationfragile}
Amirata Ghorbani, Abubakar Abid, and James Zou.
\newblock Interpretation of neural networks is fragile.
\newblock In \emph{Proceedings of the AAAI Conference on Artificial
  Intelligence}, volume~33, pp.\  3681--3688, 2019.

\bibitem[Girardi et~al.(2018)Girardi, Ji, Nguyen, Hollenstein, Ivankay, Kuhn,
  Marchiori, and Zhang]{aida}
Ivan Girardi, Pengfei Ji, An-phi Nguyen, Nora Hollenstein, Adam Ivankay, Lorenz
  Kuhn, Chiara Marchiori, and Ce~Zhang.
\newblock Patient risk assessment and warning symptom detection using deep
  attention-based neural networks.
\newblock \emph{arXiv preprint arXiv:1809.10804}, 2018.

\bibitem[Goodfellow et~al.(2016)Goodfellow, Bengio, Courville, and
  Bengio]{deeplearningbook}
Ian Goodfellow, Yoshua Bengio, Aaron Courville, and Yoshua Bengio.
\newblock \emph{Deep learning}, volume~1.
\newblock MIT Press, 2016.

\bibitem[Goodfellow et~al.(2014)Goodfellow, Shlens, and
  Szegedy]{adversarialattacks}
Ian~J Goodfellow, Jonathon Shlens, and Christian Szegedy.
\newblock Explaining and harnessing adversarial examples.
\newblock \emph{arXiv preprint arXiv:1412.6572}, 2014.

\bibitem[Hendricks et~al.(2018)Hendricks, Hu, Darrell, and
  Akata]{counterfactuals}
Lisa~Anne Hendricks, Ronghang Hu, Trevor Darrell, and Zeynep Akata.
\newblock Generating counterfactual explanations with natural language.
\newblock \emph{arXiv preprint arXiv:1806.09809}, 2018.

\bibitem[Honnibal et~al.(2020)Honnibal, Montani, Van~Landeghem, and
  Boyd]{spacy}
Matthew Honnibal, Ines Montani, Sofie Van~Landeghem, and Adriane Boyd.
\newblock {spaCy: Industrial-strength Natural Language Processing in Python},
  2020.
\newblock URL \url{https://doi.org/10.5281/zenodo.1212303}.

\bibitem[Ivankay et~al.(2020)Ivankay, Girardi, Marchiori, and Frossard]{far}
Adam Ivankay, Ivan Girardi, Chiara Marchiori, and Pascal Frossard.
\newblock Far: A general framework for attributional robustness.
\newblock \emph{arXiv preprint arXiv:2010.07393}, 2020.

\bibitem[Jacovi \& Goldberg(2020)Jacovi and Goldberg]{faithfulness}
Alon Jacovi and Yoav Goldberg.
\newblock Towards faithfully interpretable nlp systems: How should we define
  and evaluate faithfulness?
\newblock \emph{arXiv preprint arXiv:2004.03685}, 2020.

\bibitem[Jain \& Wallace(2019)Jain and Wallace]{attentionnotexplanation}
Sarthak Jain and Byron~C Wallace.
\newblock Attention is not explanation.
\newblock \emph{arXiv preprint arXiv:1902.10186}, 2019.

\bibitem[Jin et~al.(2019)Jin, Jin, Zhou, and Szolovits]{isbertrobust}
Di~Jin, Zhijing Jin, Joey~Tianyi Zhou, and Peter Szolovits.
\newblock Is bert really robust? natural language attack on text classification
  and entailment.
\newblock \emph{arXiv preprint arXiv:1907.11932}, 2, 2019.

\bibitem[Kendall(1938)]{kendall}
Maurice~G Kendall.
\newblock A new measure of rank correlation.
\newblock \emph{Biometrika}, 30\penalty0 (1/2):\penalty0 81--93, 1938.

\bibitem[Keselj(2009)]{perplexity}
Vlado Keselj.
\newblock Speech and language processing daniel jurafsky and james h. martin
  (stanford university and university of colorado at boulder) pearson prentice
  hall, 2009, xxxi+ 988 pp; hardbound, isbn 978-0-13-187321-6, \$115.00, 2009.

\bibitem[Kokhlikyan et~al.(2020)Kokhlikyan, Miglani, Martin, Wang, Alsallakh,
  Reynolds, Melnikov, Kliushkina, Araya, Yan, et~al.]{captum}
Narine Kokhlikyan, Vivek Miglani, Miguel Martin, Edward Wang, Bilal Alsallakh,
  Jonathan Reynolds, Alexander Melnikov, Natalia Kliushkina, Carlos Araya, Siqi
  Yan, et~al.
\newblock Captum: A unified and generic model interpretability library for
  pytorch.
\newblock \emph{arXiv preprint arXiv:2009.07896}, 2020.

\bibitem[La~Malfa et~al.(2021)La~Malfa, Zbrzezny, Michelmore, Paoletti, and
  Kwiatkowska]{guaranteednlprobustness}
Emanuele La~Malfa, Agnieszka Zbrzezny, Rhiannon Michelmore, Nicola Paoletti,
  and Marta Kwiatkowska.
\newblock On guaranteed optimal robust explanations for nlp models.
\newblock \emph{arXiv preprint arXiv:2105.03640}, 2021.

\bibitem[Li et~al.(2020)Li, Ma, Guo, Xue, and Qiu]{bertattack}
Linyang Li, Ruotian Ma, Qipeng Guo, Xiangyang Xue, and Xipeng Qiu.
\newblock Bert-attack: Adversarial attack against bert using bert.
\newblock \emph{arXiv preprint arXiv:2004.09984}, 2020.

\bibitem[Liu et~al.(2019)Liu, Ott, Goyal, Du, Joshi, Chen, Levy, Lewis,
  Zettlemoyer, and Stoyanov]{roberta}
Yinhan Liu, Myle Ott, Naman Goyal, Jingfei Du, Mandar Joshi, Danqi Chen, Omer
  Levy, Mike Lewis, Luke Zettlemoyer, and Veselin Stoyanov.
\newblock Roberta: A robustly optimized bert pretraining approach.
\newblock \emph{arXiv preprint arXiv:1907.11692}, 2019.

\bibitem[Maas et~al.(2011)Maas, Daly, Pham, Huang, Ng, and Potts]{imdb}
Andrew Maas, Raymond~E Daly, Peter~T Pham, Dan Huang, Andrew~Y Ng, and
  Christopher Potts.
\newblock Learning word vectors for sentiment analysis.
\newblock In \emph{Proceedings of the 49th annual meeting of the association
  for computational linguistics: Human language technologies}, pp.\  142--150,
  2011.

\bibitem[Madry et~al.(2017)Madry, Makelov, Schmidt, Tsipras, and
  Vladu]{advtraining}
Aleksander Madry, Aleksandar Makelov, Ludwig Schmidt, Dimitris Tsipras, and
  Adrian Vladu.
\newblock Towards deep learning models resistant to adversarial attacks.
\newblock \emph{arXiv preprint arXiv:1706.06083}, 2017.

\bibitem[Modas et~al.(2019)Modas, Moosavi-Dezfooli, and Frossard]{sparsefool}
Apostolos Modas, Seyed-Mohsen Moosavi-Dezfooli, and Pascal Frossard.
\newblock Sparsefool: a few pixels make a big difference.
\newblock In \emph{Proceedings of the IEEE Conference on Computer Vision and
  Pattern Recognition}, pp.\  9087--9096, 2019.

\bibitem[Moosavi-Dezfooli et~al.(2016)Moosavi-Dezfooli, Fawzi, and
  Frossard]{deepfool}
Seyed-Mohsen Moosavi-Dezfooli, Alhussein Fawzi, and Pascal Frossard.
\newblock Deepfool: a simple and accurate method to fool deep neural networks.
\newblock In \emph{Proceedings of the IEEE conference on computer vision and
  pattern recognition}, pp.\  2574--2582, 2016.

\bibitem[Moosavi-Dezfooli et~al.(2017)Moosavi-Dezfooli, Fawzi, Fawzi, and
  Frossard]{universalimage}
Seyed-Mohsen Moosavi-Dezfooli, Alhussein Fawzi, Omar Fawzi, and Pascal
  Frossard.
\newblock Universal adversarial perturbations.
\newblock In \emph{Proceedings of the IEEE conference on computer vision and
  pattern recognition}, pp.\  1765--1773, 2017.

\bibitem[Moosavi-Dezfooli et~al.(2019{\natexlab{a}})Moosavi-Dezfooli, Fawzi,
  Uesato, and Frossard]{curvatureregularization}
Seyed-Mohsen Moosavi-Dezfooli, Alhussein Fawzi, Jonathan Uesato, and Pascal
  Frossard.
\newblock Robustness via curvature regularization, and vice versa.
\newblock In \emph{Proceedings of the IEEE/CVF Conference on Computer Vision
  and Pattern Recognition}, pp.\  9078--9086, 2019{\natexlab{a}}.

\bibitem[Moosavi-Dezfooli et~al.(2019{\natexlab{b}})Moosavi-Dezfooli, Fawzi,
  Uesato, and Frossard]{robustnesscurvature}
Seyed-Mohsen Moosavi-Dezfooli, Alhussein Fawzi, Jonathan Uesato, and Pascal
  Frossard.
\newblock Robustness via curvature regularization, and vice versa.
\newblock In \emph{Proceedings of the IEEE Conference on Computer Vision and
  Pattern Recognition}, pp.\  9078--9086, 2019{\natexlab{b}}.

\bibitem[Mrk{\v{s}}i{\'c} et~al.(2016)Mrk{\v{s}}i{\'c}, S{\'e}aghdha, Thomson,
  Ga{\v{s}}i{\'c}, Rojas-Barahona, Su, Vandyke, Wen, and Young]{counterfitted}
Nikola Mrk{\v{s}}i{\'c}, Diarmuid~O S{\'e}aghdha, Blaise Thomson, Milica
  Ga{\v{s}}i{\'c}, Lina Rojas-Barahona, Pei-Hao Su, David Vandyke, Tsung-Hsien
  Wen, and Steve Young.
\newblock Counter-fitting word vectors to linguistic constraints.
\newblock \emph{arXiv preprint arXiv:1603.00892}, 2016.

\bibitem[Myers \& Sirois(2004)Myers and Sirois]{spearman}
Leann Myers and Maria~J Sirois.
\newblock Spearman correlation coefficients, differences between.
\newblock \emph{Encyclopedia of statistical sciences}, 12, 2004.

\bibitem[Paszke et~al.(2019)Paszke, Gross, Massa, Lerer, Bradbury, Chanan,
  Killeen, Lin, Gimelshein, Antiga, et~al.]{pytorch}
Adam Paszke, Sam Gross, Francisco Massa, Adam Lerer, James Bradbury, Gregory
  Chanan, Trevor Killeen, Zeming Lin, Natalia Gimelshein, Luca Antiga, et~al.
\newblock Pytorch: An imperative style, high-performance deep learning library.
\newblock In \emph{Advances in Neural Information Processing Systems}, pp.\
  8026--8037, 2019.

\bibitem[Pearson(1895)]{pcc}
Karl Pearson.
\newblock Notes on regression and inheritance in the case of two parents.
\newblock \emph{Proceedings of the Royal Society of London}, 58\penalty0
  (347-352):\penalty0 240--242, 1895.

\bibitem[Pennington et~al.(2014)Pennington, Socher, and Manning]{glove}
Jeffrey Pennington, Richard Socher, and Christopher~D Manning.
\newblock Glove: Global vectors for word representation.
\newblock In \emph{Proceedings of the 2014 conference on empirical methods in
  natural language processing (EMNLP)}, pp.\  1532--1543, 2014.

\bibitem[Radford et~al.(2019)Radford, Wu, Child, Luan, Amodei, Sutskever,
  et~al.]{gpt2}
Alec Radford, Jeffrey Wu, Rewon Child, David Luan, Dario Amodei, Ilya
  Sutskever, et~al.
\newblock Language models are unsupervised multitask learners.
\newblock \emph{OpenAI blog}, 1\penalty0 (8):\penalty0 9, 2019.

\bibitem[Serrano \& Smith(2019)Serrano and Smith]{attentioninterpretable}
Sofia Serrano and Noah~A Smith.
\newblock Is attention interpretable?
\newblock \emph{arXiv preprint arXiv:1906.03731}, 2019.

\bibitem[Simonyan et~al.(2013)Simonyan, Vedaldi, and Zisserman]{saliencymap}
Karen Simonyan, Andrea Vedaldi, and Andrew Zisserman.
\newblock Deep inside convolutional networks: Visualising image classification
  models and saliency maps.
\newblock \emph{arXiv preprint arXiv:1312.6034}, 2013.

\bibitem[Singh et~al.(2019)Singh, Kumari, Mangla, Sinha, Balasubramanian, and
  Krishnamurthy]{alignment}
Mayank Singh, Nupur Kumari, Puneet Mangla, Abhishek Sinha, Vineeth~N
  Balasubramanian, and Balaji Krishnamurthy.
\newblock On the benefits of attributional robustness.
\newblock \emph{arXiv preprint arXiv:1911.13073}, 2019.

\bibitem[Sinha et~al.(2021)Sinha, Chen, Sekhon, Ji, and Qi]{devilswork}
Sanchit Sinha, Hanjie Chen, Arshdeep Sekhon, Yangfeng Ji, and Yanjun Qi.
\newblock Perturbing inputs for fragile interpretations in deep natural
  language processing.
\newblock \emph{arXiv preprint arXiv:2108.04990}, 2021.

\bibitem[Sun et~al.(2020)Sun, Hashimoto, Yin, Asai, Li, Yu, and Xiong]{advbert}
Lichao Sun, Kazuma Hashimoto, Wenpeng Yin, Akari Asai, Jia Li, Philip Yu, and
  Caiming Xiong.
\newblock Adv-bert: Bert is not robust on misspellings! generating nature
  adversarial samples on bert.
\newblock \emph{arXiv preprint arXiv:2003.04985}, 2020.

\bibitem[Sundararajan et~al.(2017)Sundararajan, Taly, and
  Yan]{integratedgradients}
Mukund Sundararajan, Ankur Taly, and Qiqi Yan.
\newblock Axiomatic attribution for deep networks.
\newblock In \emph{Proceedings of the 34th International Conference on Machine
  Learning}, volume~70, pp.\  3319--3328, 2017.

\bibitem[Vig(2019)]{bertviz}
Jesse Vig.
\newblock Bertviz: A tool for visualizing multihead self-attention in the bert
  model.
\newblock In \emph{ICLR Workshop: Debugging Machine Learning Models}, 2019.

\bibitem[Wiegreffe \& Pinter(2019)Wiegreffe and
  Pinter]{attentionnotnotexplanation}
Sarah Wiegreffe and Yuval Pinter.
\newblock Attention is not not explanation.
\newblock \emph{arXiv preprint arXiv:1908.04626}, 2019.

\bibitem[Wolf et~al.(2020)Wolf, Debut, Sanh, Chaumond, Delangue, Moi, Cistac,
  Rault, Louf, Funtowicz, Davison, Shleifer, von Platen, Ma, Jernite, Plu, Xu,
  Scao, Gugger, Drame, Lhoest, and Rush]{huggingface}
Thomas Wolf, Lysandre Debut, Victor Sanh, Julien Chaumond, Clement Delangue,
  Anthony Moi, Pierric Cistac, Tim Rault, Rémi Louf, Morgan Funtowicz, Joe
  Davison, Sam Shleifer, Patrick von Platen, Clara Ma, Yacine Jernite, Julien
  Plu, Canwen Xu, Teven~Le Scao, Sylvain Gugger, Mariama Drame, Quentin Lhoest,
  and Alexander~M. Rush.
\newblock Transformers: State-of-the-art natural language processing.
\newblock In \emph{Proceedings of the 2020 Conference on Empirical Methods in
  Natural Language Processing: System Demonstrations}, pp.\  38--45, Online,
  October 2020. Association for Computational Linguistics.
\newblock URL \url{https://www.aclweb.org/anthology/2020.emnlp-demos.6}.

\bibitem[Yang et~al.(2020)Yang, Chen, Hsieh, Wang, and Jordan]{greedygumbel}
Puyudi Yang, Jianbo Chen, Cho-Jui Hsieh, Jane-Ling Wang, and Michael~I Jordan.
\newblock Greedy attack and gumbel attack: Generating adversarial examples for
  discrete data.
\newblock \emph{J. Mach. Learn. Res.}, 21\penalty0 (43):\penalty0 1--36, 2020.

\bibitem[Yang et~al.(2019)Yang, Dai, Yang, Carbonell, Salakhutdinov, and
  Le]{xlnet}
Zhilin Yang, Zihang Dai, Yiming Yang, Jaime Carbonell, Russ~R Salakhutdinov,
  and Quoc~V Le.
\newblock Xlnet: Generalized autoregressive pretraining for language
  understanding.
\newblock \emph{Advances in neural information processing systems}, 32, 2019.

\bibitem[Zeiler \& Fergus(2014)Zeiler and Fergus]{occlusion}
Matthew Zeiler and Rob Fergus.
\newblock Visualizing and understanding convolutional networks.
\newblock In \emph{European Conference on Computer Vision}, pp.\  818--833.
  Springer, 2014.

\bibitem[Zhang et~al.(2015)Zhang, Zhao, and LeCun]{agnews_mr}
Xiang Zhang, Junbo~Jake Zhao, and Yann LeCun.
\newblock Character-level convolutional networks for text classification.
\newblock In \emph{NIPS}, 2015.

\end{thebibliography}
\bibliographystyle{iclr2022_conference}



\end{document}